\pdfoutput=1

\documentclass[11pt]{article}

\usepackage[final]{acl}

\usepackage{times}
\usepackage{latexsym}
\usepackage{colortbl} 
\usepackage{xcolor} 
\usepackage{makecell} 
\usepackage[T1]{fontenc}

\usepackage[utf8]{inputenc}
\usepackage{amsmath}
\usepackage{amssymb}
\usepackage{mathtools}
\usepackage{amsthm}
\usepackage{booktabs}
\usepackage{microtype}

\usepackage{inconsolata}

\usepackage{graphicx}
\definecolor{fed867}{rgb}{1.0, 0.85, 0.5}
\definecolor{fed867}{rgb}{1.0, 0.949, 0.8}

\title{Following the Autoregressive Nature of LLM Embeddings \\via Compression and Alignment}

\author{Jingcheng Deng$^{1,2\ *}$, Zhongtao Jiang$^{3}$\thanks{\ \ Equal Contributions},Liang Pang$^{1}$\thanks{\ \ Corresponding Author}, Zihao Wei$^{1,2}$, \\
\textbf{Liwei Chen$^{3}$,Kun Xu, Yang Song, Huawei Shen$^{1,2}$, Xueqi Cheng$^{1,2}$}\\
  $^{1}$Key Laboratory of AI Safety, Institute of Computing Technology,\\ Chinese Academy of Sciences \\
  $^{2}$ University of Chinese Academy of Sciences \\
  $^{3}$ Kuaishou Technology\\
\texttt{\{dengjingcheng23s, pangliang\}@ict.ac.cn} \\}

\begin{document}
\maketitle
\begin{abstract}
A new trend uses LLMs as dense text encoders via contrastive learning. However, since LLM embeddings predict the probability distribution of the next token, they are inherently generative and distributive, conflicting with contrastive learning, which requires embeddings to capture full-text semantics and align via cosine similarity. This discrepancy hinders the full utilization of LLMs' pre-training capabilities, resulting in inefficient learning. In response to this issue, we propose AutoRegEmbed, a new contrastive learning method built on embedding conditional probability distributions, which integrates two core tasks: information compression and conditional distribution alignment. The information compression task encodes text into the embedding space, ensuring that the embedding vectors capture global semantics. The conditional distribution alignment task focuses on aligning text embeddings with positive samples embeddings by leveraging the conditional distribution of embeddings while simultaneously reducing the likelihood of generating negative samples from text embeddings, thereby achieving embedding alignment and uniformity. Experimental results demonstrate that our method significantly outperforms traditional contrastive learning approaches and achieves performance comparable to state-of-the-art models when using the same amount of data. Our code is available at \url{https://github.com/TrustedLLM/AutoRegEmbed}
\end{abstract}

\section{Introduction}
Text embeddings, which represent the semantic content of natural language text as vectors, are extensively utilized in domains such as information retrieval \citep{DBLP:conf/sigir/XiaXLGC15}, semantic similarity assessment, retrieval-augmented generation (RAG) \citep{DBLP:conf/iclr/KhandelwalLJZL20, DBLP:conf/naacl/ShiMYS0LZY24, DBLP:conf/emnlp/DengPSC23, DBLP:journals/corr/abs-2402-10612,DBLP:conf/acl/XuPYMSCZ24, DBLP:conf/iclr/XuPSC25}, LLMs-based agents \citep{DBLP:conf/ijcai/ChenDSZS00024,DBLP:conf/www/XuPSCC24} and data attribution \citep{DBLP:conf/dsaa/BeigiTMCS024}. Traditional text embedding models typically employ transformer-based architectures with encoder-only designs, including examples like Bert \citep{DBLP:conf/naacl/DevlinCLT19}, DeBERTa \citep{DBLP:conf/iclr/HeLGC21} and MPNet \citep{DBLP:conf/nips/Song0QLL20}, and are trained using contrastive learning.
\begin{figure}[t]
  \includegraphics[width=1.\columnwidth]{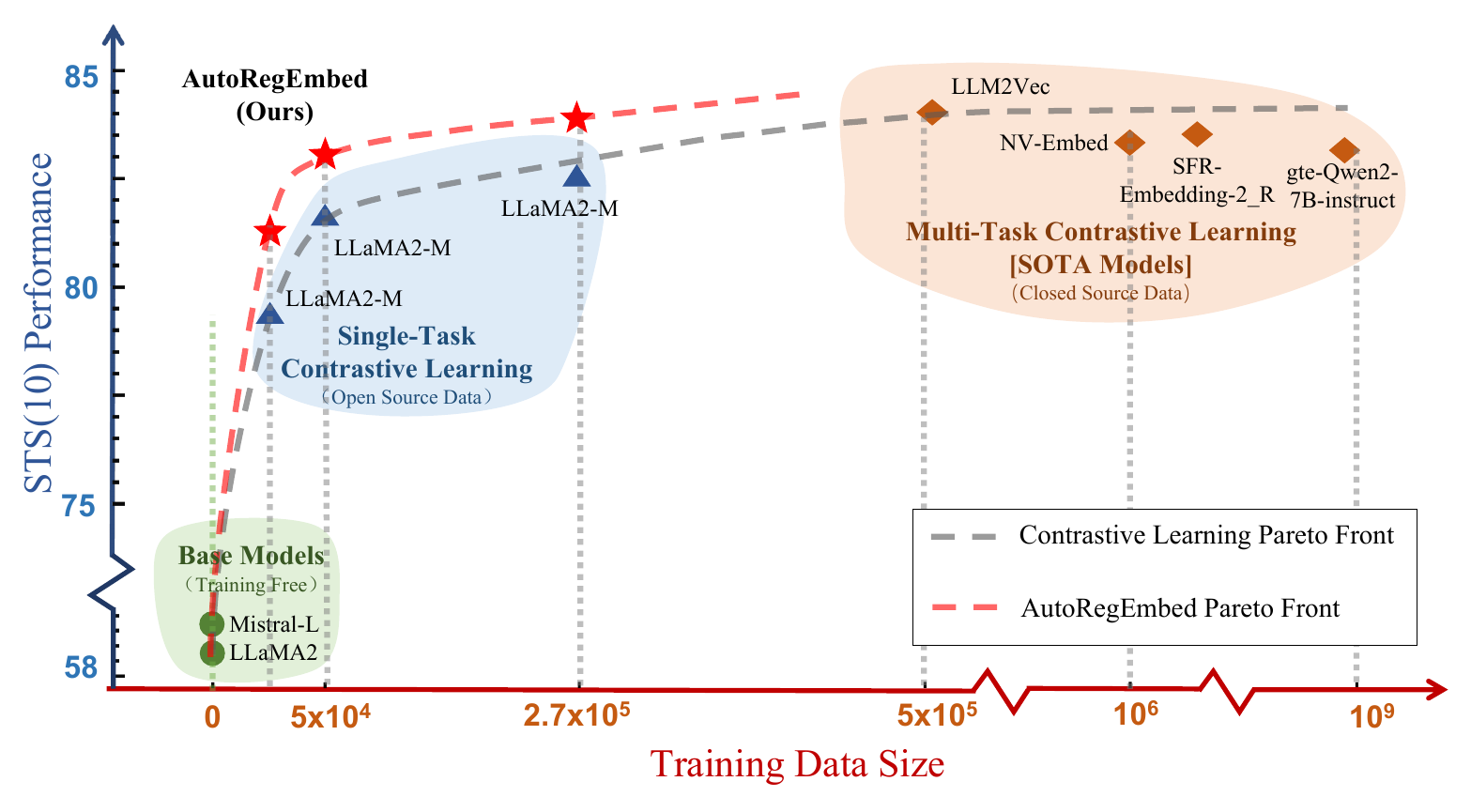}
  \caption{Comparison of pareto front between AutoRegEmbed and other methods. The horizontal axis represents the number of training samples, while the vertical axis indicates the average performance across 10 STS datasets. The upper left corner represents the region with the highest learning efficiency.}
  \label{fig:comparison}
  \vspace{-0.5cm}
\end{figure}

After extensive pre-training on a large-scale corpus, LLMs have outperformed previous encoder-only small models \citep{DBLP:conf/emnlp/DengDGJP22} and demonstrated strong adaptability across diverse downstream tasks \citep{zhao2024large,wang-etal-2024-gpt,zhao2025roleplayparadoxlargelanguage,duan-etal-2025-related, liu2025forewarnedforearmedpresynthesizingjailbreaklike}. Recently, contrastive learning has been directly applied to decoder-only LLMs, which are trained to generate embedding vectors based on task-specific instructions, enabling adaptability to various embedding scenarios \cite{ DBLP:journals/corr/abs-2405-17428, DBLP:journals/corr/abs-2404-05961}. Despite initial advancements, training a high-performance 7B-scale text embedding model using this approach remains highly resource-intensive. It typically requires millions of triplets \cite{DBLP:conf/acl/WangYHYMW24, DBLP:journals/corr/abs-2409-15700, DBLP:journals/corr/abs-2308-03281} and substantial computational power, including thousands of hours on an A100 80GB GPUs \cite{DBLP:journals/corr/abs-2402-09906, DBLP:conf/sigir/MaWYWL24}, even with the application of Parameter-Efficient Fine-Tuning (PEFT) \cite{DBLP:conf/iclr/HuSWALWWC22, DBLP:conf/iclr/Dao24}. The high resource consumption might reasonably be attributed to the inability of the \textit{discriminative} contrastive learning method to fully harness the capabilities of \textit{generative} LLMs \cite{DBLP:conf/acl/Li0XSL24}. Firstly, the constraint of unidirectional attention in LLMs leads to the aggregation of information in the hidden state of the output layer corresponding to the final token. However, as LLMs are optimized for next-token prediction, this hidden state can only represent the semantics of the next token (local) rather than the semantics of the input text itself (global). Consequently, employing this hidden state directly in contrastive learning requires additional training time and computational cost to transition from a localized to a more global semantic representation. Secondly, the hidden state in LLMs is used to generate the probability distribution of the next token, whereas contrastive learning optimizes the cosine distance between the hidden states of different texts. This divergence in optimization objectives introduces additional training costs. This raises an important question: \textit{Is it feasible to develop a method that follows the auto-regressive nature while generating high-quality text embeddings and significantly reducing resource requirements?}

We formalize three key requirements to address this problem. Firstly, embeddings should capture global semantics rather than focusing solely on next-token semantics. Secondly, they must follow alignment and uniformity principles \cite{DBLP:conf/icml/0001I20}. Finally, the transformation from the original embedding to one that meets these criteria should follow an autoregressive nature. To this end, we propose AutoRegEmbed, which encompasses two tasks: information compression and conditional distribution alignment.

The information compression task is inspired by the concept of context compression \cite{DBLP:conf/emnlp/ChevalierWAC23, DBLP:conf/iclr/00010WWCW24, DBLP:conf/nips/Mu0G23}, which addresses the limitations of context window length and the high computational cost faced by LLMs when processing long texts. Specifically, we encode the context and instructions into a set of compressed variables, which are then passed to a decoder with the same architecture but frozen parameters, forcing it to reconstruct the corresponding target. By restricting the decoder to rely solely on the compressed variables—without access to the original context or instructions—we introduce an information bottleneck. This ensures that the compressed variables effectively capture the global semantics of the instructions and context.

The conditional distribution alignment task draws inspiration from traditional contrastive learning and LLM alignment techniques \cite{DBLP:journals/corr/abs-2407-16216}. We begin by treating the compressed vectors as embeddings of their corresponding inputs. Then, we adopt the structure of the InfoNCE \citep{DBLP:journals/corr/abs-1807-03748} loss function, but redefine the similarity metric. Simply put, we align the distance between the conditional probability distributions of text and positive sample embeddings while increasing the likelihood of text embeddings generating positive samples and decreasing the likelihood of generating negative samples. This approach promotes the alignment and uniformity of compressed variables while maintaining the autoregressive nature.

Experimental results demonstrate that AutoRegEmbed outperforms traditional contrastive learning methods while utilizing the same computational resources, making it a highly efficient and scalable solution. Remarkably, even with a limited number of training samples, AutoRegEmbed achieves performance on par with the current state-of-the-art (SOTA) models, showcasing its superior ability to learn robust and generalizable representations from scarce data. As shown in Figure~\ref{fig:comparison}, the Pareto frontier of AutoRegEmbed consistently outperforms traditional contrastive learning methods, demonstrating a more optimal trade-off between computational efficiency and performance. This indicates that AutoRegEmbed achieves superior representation learning while maintaining a balanced resource utilization.

\section{Related Works}
Text embedding is a technique that maps text data into a numerical vector space, capturing both semantic and contextual features of the text. Research in this area can be divided into three categories based on the underlying model: Early models, LLMs with fine-tuning, and LLMs without fine-tuning.

\paragraph{Early Models} Early approaches
include SentenceBERT \cite{DBLP:conf/emnlp/ReimersG19} (supervised) and SimCSE \cite{DBLP:conf/emnlp/GaoYC21} (unsupervised), which leverage contrastive learning to generate high-quality text embeddings using small encoder-only models. 
Inspired by instruction fine-tuning, recent research \cite{DBLP:conf/acl/SuSKWHOYSZ023, DBLP:conf/acl/AsaiSL0I0HY23} has shifted toward using text paired with instructions to enhance the generalization and transferability of text embeddings in complex scenarios.  At this stage, several studies have investigated the use of generative tasks to enhance text embeddings. For example, coCondenser \citep{DBLP:conf/acl/GaoC22} proposes a pre-training strategy that improves the expressiveness of the [CLS] token using unsupervised corpora. PaSeR \citep{DBLP:conf/emnlp/WuZ22} and RetroMAE \citep{DBLP:conf/emnlp/XiaoLSC22} adopt encoder-decoder architectures: PaSeR emphasizes the reconstruction of key phrases, while RetroMAE employs a masked-token objective to recover the original sentence.
However, these approaches do not incorporate contrastive supervision signals into their generative objectives. Moreover, they are evaluated only on smaller-scale models, making them fundamentally different from our method in both design philosophy and scaling strategy.


\paragraph{LLMs with Fine-Tuning}
Many studies have focused on transforming LLMs into text embedding models through contrastive learning fine-tuning. RepLLaMA \citep{DBLP:conf/sigir/MaWYWL24}, for example, follows the DPR \citep{DBLP:conf/emnlp/KarpukhinOMLWEC20} pipeline, using the hidden state of the last token generated by LLaMA as a text embedding vector and applying contrastive learning fine-tuning. Recognizing that the unidirectional attention mechanism in LLMs may limit text embedding quality, LLM2Vec \citep{DBLP:journals/corr/abs-2404-05961} introduces a bidirectional attention mechanism combined with average pooling to enhance embedding quality. NV-Embed \citep{DBLP:journals/corr/abs-2405-17428} takes this further by incorporating an additional Latent Attention Layer to generate pooled embeddings. bge-en-icl \cite{DBLP:journals/corr/abs-2409-15700} suggests that retaining the original framework of LLMs and leveraging in-context learning is the optimal approach for generating text embeddings. Some studies \cite{DBLP:conf/acl/WangYHYMW24} even use synthetic data generated by LLMs, rather than real-world data, for fine-tuning and achieve competitive performance on the MTEB leaderboard \cite{DBLP:conf/eacl/MuennighoffTMR23}. However, these approaches often overlook the fundamental differences between language modeling and contrastive learning, failing to fully leverage the potential of LLMs. More closely related to our work is Llama2Vec \cite{DBLP:conf/acl/Li0XSL24}, which proposes two pretext tasks to enable unsupervised adaptation of LLMs, followed by contrastive learning fine-tuning to achieve better performance. In contrast, our approach achieves strong results without any need for traditional contrastive learning fine-tuning (cosine-based), as our task fully exploits the inherent potential of LLMs.

\paragraph{LLMs without Fine-Tuning}
Several studies have explored methods to transform LLMs into text encoders without fine-tuning. \cite{DBLP:conf/iclr/LiuTAPZS24} proposed using possible trajectory distributions as text representations, achieving effectiveness but at a high computational cost. \citep{DBLP:journals/corr/abs-2402-15449} introduced echo embeddings by repeatedly feeding text into autoregressive models, addressing architectural limitations but doubling computational requirements. Other methods focus on prompt adjustments to produce meaningful embeddings. PromptEOL \citep{DBLP:conf/emnlp/JiangHLWZ24} introduced a One-Word Limitation prompt to improve embedding performance, while MetaEOL \citep{DBLP:conf/acl/LeiW00CTY24} extended this idea by using eight different prompt types to generate multi-view embeddings. GenEOL \citep{DBLP:journals/corr/abs-2410-14635} leveraged LLMs to create various sentence transformations that retain their meaning, aggregating the resulting embeddings to enhance the overall sentence representation. Meanwhile, PromptReps \cite{DBLP:conf/emnlp/ZhuangMKLZ24} developed a hybrid document retrieval framework leveraging prompts to address challenges in information retrieval tasks. Despite these innovations, these approaches either perform poorly or require multiple inferences to achieve good results. By contrast, our method surpasses these methods with minimal training costs.

\section{Method}
In this section, we first introduce the preliminary information about the task of text embedding with instructions. We then discuss the information compression, which transitions LLM embeddings from local semantics to global semantics, followed by the conditional distribution alignment, which optimizes the conditional probability distribution of embeddings to ensure alignment and uniformity.
\begin{figure*}[t]
  \includegraphics[width=1.\textwidth]{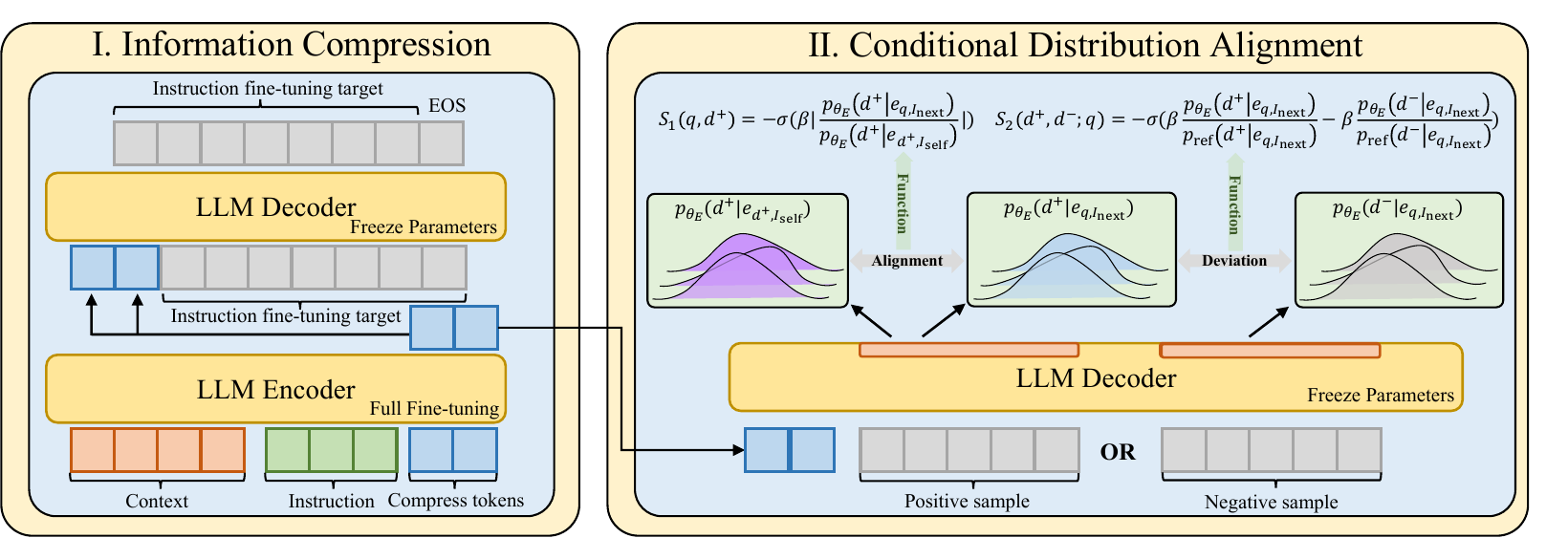}
  \caption{Overall framework of AutoRegEmbed. Firstly, we perform the information compression task to inject key information from the context and instruction into the compressed tokens. Then, we optimize the conditional probability distribution of these tokens to align the distributions of $e_{q,I_{\mathrm{next}}}$ and $e_{d^+,I_{\mathrm{self}}}$ as closely as possible through $S_1(q,d^+)$, while increasing the probability of $e_{q,I_{\mathrm{next}}}$ generating positive samples and reducing the probability of $e_{q,I_{\mathrm{next}}}$ generating negative samples through $S_2(d^+,d^-;q)$. Encoder and decoder share a structure.} 
  \label{fig:overview}
\end{figure*}
\subsection{Preliminary}
\label{sec:preliminary}
Text embeddings with instructions can adapt to various downstream tasks. Formally, given a large collection 
$D=\{d_1,d_2,\dots,d_N\}$ containing $N$ documents, as well as a text $q$ and an instruction $t$, the embedding $e_{q,t}=E(q,t)$ generated from $q$ and $t$ can match documents $d \in D$ that are relevant to $q$, according to $t$, where $E$ represents the text encoder. Thus, by simply changing the instruction $t$, the relevance measure can be adapted to different downstream tasks. For example, for dense retrieval tasks, the instruction might be ``\textit{find documents that can answer this question,}'' while for semantic similarity tasks, the instruction could be ``\textit{find sentences that are semantically similar to this text}''. Numerous studies have explored various embedding techniques and instruction diversities. Our goal is to identify a simple yet effective way to enable LLMs to generate high-quality embeddings directly from autoregressive framework.
\subsection{Information Compression: from Discriminative to Generative Embeddings} 
In this section, we first explain the motivation for transitioning from discriminative embeddings to generative embeddings, followed by a formal definition of the information compression task.

In decoder-based LLMs, embeddings are typically generated by extracting the hidden state of the final token in the input sequence. However, this approach primarily captures the semantics of the first output token rather than encoding the global semantics of the entire input. Various pooling techniques, such as average pooling and attention pooling, have been explored to mitigate this limitation, yet they introduce their own challenges. The average pooling method, which computes the mean of all token hidden states, does not necessarily encapsulate global semantics but instead serves as a mechanism for "convexity preservation" \cite{DBLP:conf/emnlp/LiZHWYL20}. Conversely, attention pooling modifies the attention mechanism or introduces additional parameters, thereby altering the original architecture of LLMs. Such modifications deviate from the model’s pre-training design and can lead to unintended consequences, as prior studies \cite{DBLP:journals/corr/abs-2409-15700} indicate that maintaining the original LLM framework often yields optimal performance. To enable LLMs to generate embeddings that represent global semantics, we introduce an information compression task. This task compels LLMs to reconstruct the original target using a compressed embedding derived from the input text. Given that this compressed embedding models the conditional probability distribution of the target, we designate it as the generative embedding to contrast it with the discriminative embedding produced by conventional pooling approaches.

The information compression task is inspired by the concept of context compression. Specifically, we append $k$ compressed tokens $c=(c_1,\dots,c_k)$, where $k<<n+m$, to the text $q=(q_1,\dots,q_n)$ and instruction $t=(t_1,\dots,t_m)$, with $n$ and $m$ representing their respective token lengths. This combined $(q,t,c)$ is then fed into an encoder $E$ to generate the embedding $e_c=(e_{c_1},\dots,e_{c_k})$. As mentioned earlier, we expect the embedding $e_c$ to capture the global semantics of the text $q$ and the instruction $t$.  To achieve this, we input $e_c$ into a frozen decoder $D$, which shares the same architecture, and force it to generate the most relevant document $d$. The optimization objective for this task can be expressed as: 
\begin{equation*}
\begin{aligned}
\label{eq1}
    &\mathcal{L}_{\mathrm{IC}}=\underset{e_{c_1},\dots,e_{c_k}}{\mathrm{max}} P(d|e_{c_1},\dots,e_{c_k};\theta_{D})\\
    &=\underset{\Theta_{E}}{\mathrm{max}}\, P(d|c_1\dots c_k,t_1\dots t_m,q_1 \dots q_n;\theta_{E},\theta_{D}),
\end{aligned}
\end{equation*}
where $\theta_{E}$ and $\theta_{D}$ denote the parameters of $E$ and $D$, respectively.

\subsection{Conditional Distribution Alignment: from Data-Point to Distribution Perspective}
After addressing the global semantic representation issue of the embedding vector, we also require the embedding vector to meet the criteria of alignment and uniformity. In general, we optimize these two properties asymptotically using a contrastive loss, such as InfoNCE,


\begin{equation}
\begin{aligned}
\label{eq4}
    &\mathcal{L}_{\mathrm{InfoNCE}}(f;\tau)=&\\
    &\mathbb{E}[-\mathrm{log}\frac{e^{f(q)^Tf(d^+)/\tau}}{e^{f(q)^Tf(d^+)/\tau}+\sum_i e^{f(q)^Tf(d^-_i)/\tau}}] ,
\end{aligned}
\end{equation}
where $\tau$ denotes the temperature parameter and $d^-_i$ represents the $i$-th negative sample. Clearly, Equation \ref{eq4} differs fundamentally from the generative pre-training task, as it optimizes the cosine distance between sample embeddings, aligning data points in the embedding space rather than modeling the next-token probability distribution, which is central to pre-training. So, using this loss function to optimize an LLM may not fully unlock its potential.

To address this, we propose the Conditional Distribution Alignment task to minimize this discrepancy as much as possible. The concept is straightforward: Instead of using the cosine distance between embeddings, we assess similarity based on the conditional probability distribution corresponding to each embedding. Simply put, we extend point alignment to distribution alignment. Formally, the decoder $L_D$ is a well-trained autoregressive language model with the following conditional probability distribution: 
\begin{equation*}
\label{eq5}
    p(d|e_c)=\prod\limits_{t=1}^Tp(d_t|d_{<t},e),
\end{equation*}
where $e_c=(e_{c_1},\dots,e_{c_k})$ is the embedding variables, $d=(d_1,d_2,\dots,d_T)$ represents the generated sentence, and $d_{<t}$ denotes the part of the sentence before time step $t$. Intuitively, the similarity between corresponding samples $q$ and $d$ can be measured by computing the distance between the conditional probability distributions of their embeddings, $e_q$ and $e_d$:
\begin{equation*}
\label{eq6}
    S(q,d) = \frac{1}{T}\sum_{t=1}^T \mathrm{D}(p(d_t|d_{<t},e_q), p(d_t|d_{<t},e_d)),
\end{equation*}
where $\mathrm{D}(\cdot,\cdot)$ is any function that measures the divergence between two probability distributions.
Given that the embedding distribution is instruction-dependent (see Section~\ref{sec:preliminary}), we can adopt multiple approaches to measure similarity between q and $d^+$/$d^-$." Our empirical approach uses a basic alignment strategy: aligning the probability distributions of q and d via distinct instructions ($I_{\mathrm{next}}$ and $I_{\mathrm{self}}$; see Appendix~\ref{app:instrucitons}). This increases the probability of q generating $d^+$ and decreases that of generating $d^-$. As shown in Appendix E, this simple alignment outperforms more complex alternatives. Building on the above insights and incorporating the structure of InfoNCE, we empirically derive the final loss function:
\vspace{-0.2cm}
\begin{equation}
\begin{aligned}
\label{eq7}
&\mathcal{L}_{\mathrm{CDA}}=\mathbb{E}[-\mathrm{log}\frac{e^{S_1(q,d^+)/\tau}}{e^{S_1(q,d^+)/\tau}+\sum_i e^{S_2(d^+,d^-_i;q)/\tau}}], \\
& S_1(q,d^+) = - \sigma(\beta \ |\mathrm{log} \frac{p_{\theta_{E}}(d^+|e_{q,I_{\mathrm{next}}})}{p_{\theta_{E}}(d^+|e_{d^+,I_{\mathrm{self}}})}|  ),\\
& S_2(d^+,d^-_i;q) = 
- \sigma(\beta \ \mathrm{log}\frac{p_{\theta_{E}}(d^+|e_{q,I_{\mathrm{next}}})}{p_{\mathrm{ref}}(d^+|e_{q,I_{\mathrm{next}}})}\\
&-\beta \ \mathrm{log}\frac{p_{\theta_{E}}(d^-_i|e_{q,I_{\mathrm{next}}})}{p_{\mathrm{ref}}(d^-_i|e_{q,I_{\mathrm{next}}})}),
\end{aligned}
\end{equation}
where $\tau$ and $\beta$ are temperature parameters, and $p_{\Theta_{E}}$ represents the initial model. $p_{ref}$ denotes the log probability of the model generating the target before conditional distribution alignment. We use the Sigmoid function $\sigma(\cdot)$ (see Appendix~\ref{app:analysis_sigma} for analysis) to normalize the similarity measured from the conditional probability distribution to the range [0,1]to prevent overflow during the exponential operation. $S_1$ represents the similarity function between text $q$ and the positive sample $d^+$. We define it by measuring the absolute value of the difference in the logarithmic probability of their corresponding embeddings, $e_{q,I_{\mathrm{next}}}$ and $e_{d^+,I_{\mathrm{self}}}$, generating the positive sample $d^+$. To minimize this difference, we apply the absolute value function. In addition, we then add a negative sign to ensure that the value of $S_1$ increases as the similarity between $q$ and $d^+$ increases. $S_2$ calculates the difference between the logarithmic probabilities of generating positive and negative samples for text $q$, similar to DPO \citep{DBLP:conf/nips/RafailovSMMEF23}. We amplify this difference to boost the probability of embedding $e_{q,I_{\mathrm{next}}}$ generating positive samples and decrease the probability of generating negative samples. We normalize the probabilities by dividing them by the corresponding values from the initial model to account for the length discrepancy between positive and negative samples. Note that during inference, the model still computes similarity between $e_{q,I_{\mathrm{next}}}$ and $e_{d,I_{\mathrm{self}}}$ using inner product matching, enabling seamless integration with existing embedding systems.

\begin{table*}[htb]
\centering
\normalsize
\setlength{\tabcolsep}{3pt}
\resizebox{\linewidth}{!}{%
\begin{tabular}{lrccccccccccccc}
\toprule
\textbf{Method} & \textbf{Params} & \textbf{BIOSSES} &\textbf{SICK-R}&\textbf{STS12} & \textbf{STS13} & \textbf{STS14} & \textbf{STS15} & \textbf{STS16} &\textbf{STS17}&\textbf{STS22}& \textbf{STS-B}  & \textbf{Avg.}& \textbf{Vol.}\\
\midrule
\midrule
\multicolumn{10}{l}{\it{Without Contrastive Training}}\\
\rowcolor{fed867} LLaMA2-L & 7B & 63.29 & 65.10 & 45.26 & 70.83 & 56.69 & 62.48 &      63.27 & 49.76&-7.76&60.43&60.58$_{(7)}$/56.91$_{(10)}$&0\\
\rowcolor{fed867} LLaMA2-M & 7B & 65.96 & 60.01 & 44.76 & 64.13 & 48.66 & 62.33 &      63.16 & 64.35&27.59&53.50&56.65$_{(7)}$/58.67$_{(10)}$&0\\
\rowcolor{fed867} Mistral-v0.1-L & 7B & 54.40 & 67.40 & 48.54 & 64.27 & 54.89 & 65.05 &      62.12 & 48.22&13.71&63.05&60.76$_{(7)}$/56.20$_{(10)}$&0\\
\rowcolor{fed867} Mistral-v0.1-M & 7B & 67.46 & 62.42 & 50.11 & 66.45 & 52.60 & 61.93 &      65.02 & 71.28&29.79&54.19&58.96$_{(7)}$/61.13$_{(10)}$&0\\
\midrule
{Echo-LLaMA2}& 7B
 &-&64.39& 52.40 & 72.40 & 61.24 & 72.67 & 73.51 & - & - & 65.73 &66.05$_{(7)}$/-&0\\
{Echo-LLaMA2} & 13B
 &-&70.27& 59.36 & 79.01 & 69.75 & 79.86 & 76.75  & - & - &71.31&72.33$_{(7)}$/-&0\\
{PromptEOL-LLaMA2} & 7B
 & - & 69.64 & 58.81 & 77.01 & 66.34 & 73.22 & 73.56 & - & - & 71.66  &70.03$_{(7)}$/-&0\\
{PromptEOL-Mistral} & 7B
 &-&69.47& 63.08 & 78.58 & 69.40 & 77.92 & 79.01 & - &-	 & 75.77 &73.32$_{(7)}$/-&0\\
{PromptEOL-LLaMA3} & 8B
 &-& 60.88&68.94 & 78.57 & 68.18 & 76.75 & 77.16 &  -      & - & 72.83  &71.90$_{(7)}$/-&0\\
{PromptEOL-LLaMA2} & 13B
 & -&68.23& 56.19 & 76.42 & 65.42 & 72.73 & 75.21 & - & - &67.96 &68.83$_{(7)}$/-&0\\
 {MetaEOL-LLaMA2} & 7B
 & - & 74.86 & 64.16 & 81.61 & 73.09 & 81.11 & 78.94 & - & - & 77.96  &75.96$_{(7)}$/-&0\\
 {MetaEOL-Mistral} & 7B
 &-&75.13& 64.05 & 82.35 & 71.57 & 81.36 & 79.85 & - &-	 & 78.29 &76.09$_{(7)}$/-&0\\
 {GenEOL-LLaMA2-Mistral} & 7B
 &-&78.08& 70.24 & 83.43 & 78.03 & 81.79 & 80.65 & - &-	 & 80.46 &78.95$_{(7)}$/-&0\\
 {GenEOL-LLaMA2-ChatGPT} & 7B
 &-&78.71& 70.78 & 83.28 & 77.75 & 82.10 & 80.45 & - &-	 & 79.83 &78.99$_{(7)}$/-&0\\
\midrule
\midrule
\multicolumn{10}{l}{\it{Unsupervised Contrastive Training}}\\
LLM2Vec-LLaMA2$^\clubsuit$ & 7B
& 82.41&71.77&65.39 & 79.26 & 72.98 & 82.72 & 81.02 & 86.70 & 63.47 & 78.32&75.92$_{(7)}$/76.41$_{(10)}$&\textasciitilde160,000  \\
LLM2Vec-Mistral$^\clubsuit$ & 7B
 & 83.29 & 75.55 & 67.65 & 83.90 & 76.97 & 83.80 & 81.91 & 85.58 &65.93&80.42&78.60$_{(7)}$/78.50$_{(10)}$&\textasciitilde160,000\\
\midrule
\midrule
\multicolumn{10}{l}{\it{Supervised Contrastive Training}}\\
NV-Embed$^\clubsuit$ & 7.73B
& 85.59&82.80&76.22 & 86.30 & 82.09 & 87.24 & 84.77 & 87.42 & \textbf{69.85} & 86.14&83.65$_{(7)}$/82.84$_{(10)}$ &1,054,000 \\
SFR-Embedding-2\_R$^\clubsuit$ & 7B
& \textbf{87.60}&77.01&75.67 & 82.40 & 79.93 & 85.82 & 84.50 & 88.93 & 67.10 & 83.60&81.28$_{(7)}$/81.26$_{(10)}$&\textasciitilde1,751,000  \\
gte-Qwen2-7B-instruct$^\clubsuit$ & 7.49B
& 81.37&79.16&79.53 & \textbf{88.97} & 83.87 & 88.48 & 86.49 & 88.75 & 67.16 & 86.81&84.76$_{(7)}$/83.06$_{(10)}$&\textasciitilde791,000,000  \\
LLM2Vec-LLaMA2$^\clubsuit$ & 7B
& 82.13&83.01&78.85 & 86.84 & 84.04 & 88.72 & 86.79 & \textbf{90.63} & 67.55 & \textbf{88.72}&85.28$_{(7)}$/83.73$_{(10)}$&544,000  \\
LLM2Vec-Mistral$^\clubsuit$ & 7B
 & 85.24 & 83.70 & 78.80 & 86.37 & 84.04 & 88.99 & 87.22 & 90.19 &67.68&\underline{88.65}&85.40$_{(7)}$/84.01$_{(10)}$&544,000\\
 \rowcolor{fed867} {LLaMA2-L} & 7B
 & 77.58 & 77.85 & 73.72 & 84.04 & 79.82 & 85.03 & 84.78 & 87.53&26.87&86.18&81.63$_{(7)}$/76.34$_{(10)}$&50,000\\
 \rowcolor{fed867} {LLaMA2-inbatch-L} & 7B
 & 78.81 & 82.76 & 77.70 & 85.01 & 81.82 & 88.30 & 86.12 & 90.53&20.70&87.94&84.24$_{(7)}$/77.97$_{(10)}$&50,000\\
  \rowcolor{fed867} {LLaMA2-M} & 7B
 & 75.65 & 78.92 & 74.12 & 84.17 & 80.00 & 85.63 & 83.28 &85.65&65.09&86.27&81.77$_{(7)}$/79.88$_{(10)}$&50,000\\
 \rowcolor{fed867} {LLaMA2-inbatch-M} & 7B
 & 78.09 & 83.17 & 77.10 & 82.82 & 80.53 & 87.40 & 84.43 &90.02&64.59&87.18&83.23$_{(7)}$/81.53$_{(10)}$&50,000\\
  \rowcolor{fed867} {LLaMA2-inbatch-M} & 7B
 & 77.43 & 82.26 & 77.95 & 84.90 & 82.06 & 87.22 & 86.43 &88.22&66.42&86.12&83.85$_{(7)}$/81.90$_{(10)}$&274,951\\
\midrule
\midrule
\multicolumn{10}{l}{\it{Information Compression and Conditional Distribution Alignment}}\\
{AutoRegEmbed-LLaMA2} & 7B
 & 84.65 & 81.46 & \textbf{79.98} & 86.35 & 83.33 & 89.21 & 86.91 & 87.67 & 65.90 & 86.98 & 84.89$_{(7)}$/83.24$_{(10)}$ &50,000(16,382)\\
 {AutoRegEmbed-Mistral} & 7B
 & 86.84 & 80.32 & 78.92 & 86.18 & 83.29 & 88.98 & 86.75 & 88.77 &64.53&87.24&$84.53_{(7)}$/83.18$_{(10)}$&50,000(16,382)\\
 {AutoRegEmbed-LLaMA2} & 7B
 & 85.62 & \underline{83.87} & \underline{79.60} & 87.36 & \textbf{84.29} & \underline{89.43} & \textbf{87.72} & 89.46 &\underline{67.78}&87.96&\underline{85.75}$_{(7)}$/\underline{84.31}$_{(10)}$&274,951(16,382)
 \\
 {AutoRegEmbed-Mistral} & 7B
 & \underline{87.48} & \textbf{83.90} & 79.56 & \underline{87.64} & \underline{84.11} & \textbf{89.58} & \underline{87.46} & \underline{89.87} &67.77&88.48&\textbf{85.82}$_{(7)}$/\textbf{84.59}$_{(10)}$&274,951(16,382)\\
\midrule
\bottomrule
\end{tabular}
}
\caption{Results on STS tasks (Spearman correlation scaled by 100x). The parentheses in the \textbf{Avg.} column indicate the number of datasets used to compute the average. \textbf{Vol.} denotes the number of training triplets, while the numbers in brackets indicate the instruction fine-tuning data used by AutoRegEmbed during the information compression stage. The symbol “\textasciitilde” denotes an estimated value. "\textcolor{fed867}{\rule{5mm}{2mm}}" represents our own fair baselines, and we apply a grid search to ensure optimal performance. "-L" and "-M" denote the hidden state of the last token and the average pooling of all token hidden states, respectively. The symbol $\clubsuit$ indicates that not all data are open source. Bold indicates the best result, and underline indicates the second-best (suboptimal) result.
}
\label{tab:sts}
\vspace{-0.5cm}
\end{table*}
\section{Experiments}

\subsection{Experimental Settings}
\paragraph{Evaluations} Previous studies \cite{DBLP:conf/emnlp/GaoYC21,DBLP:conf/emnlp/LiZHWYL20} highlight that a key goal of text embedding is to cluster semantically similar sentences. Following this approach, we use the MTEB \cite{DBLP:conf/eacl/MuennighoffTMR23} evaluation framework to evaluate AutoRegEmbed on ten semantic text similarity datasets, including STS12 \citep{DBLP:conf/semeval/AgirreCDG12}, STS13 \citep{DBLP:conf/starsem/AgirreCDGG13}, STS14 \citep{DBLP:conf/semeval/AgirreBCCDGGMRW14}, STS15 \citep{DBLP:conf/semeval/AgirreBCCDGGLMM15}, STS16 \citep{DBLP:conf/semeval/AgirreBCDGMRW16}, STS17 \citep{DBLP:conf/semeval/CerDALS17}, STS22 \citep{DBLP:conf/semeval/ChenZCFGGHJS22}, STS-B , BIOSSES and SICK-R. Each pair of text in the STS dataset is labeled with a similarity score ranging from 0 to 5 or 0 to 4, indicating their semantic similarity. The evaluation metric is the Spearman correlation between the similarity scores predicted by the model and the scores annotated by humans. In addition, to evaluate retrieval performance, we assess the model on MS MARCO \citep{DBLP:conf/nips/NguyenRSGTMD16}, NFCorpus \citep{DBLP:conf/ecir/BotevaGSR16}, and SCIDOCS \citep{DBLP:conf/acl/CohanFBDW20} using nDCG@10 as the evaluation metric.

\paragraph{Training} In the information compression stage, we use the training set of the instruction fine-tuning dataset PWC \citep{DBLP:conf/iclr/00010WWCW24}, which includes a diverse range of instruction types, as the training data. The original dataset contains 241,564 (context, instruction, target) samples. To reduce redundancy caused by repeated contexts, we remove duplicates, resulting in the PWC-Unique dataset with 16,382 samples as the final training data. In the conditional distribution alignment stage, we use the NLI part of the MEDI \citep{DBLP:conf/acl/WangYHYMW24} and BGE \citep{DBLP:journals/corr/abs-2402-03216} datasets as training data. The former contains 50,000 samples, while the latter consists of 274,951 samples. Each sample includes an anchor, a positive sample, and a negative sample. Unless otherwise specified, the AutoRegEmbed results presented in the experiment section are based on the NLI part of the MEDI. For the retrieval task, we train our model on the MS MARCO training set. Because of the asymmetric nature of the retrieval task, the gap between the cosine distance–based evaluation metric and the conditional probability–based optimization objective becomes pronounced. To mitigate this, we perform an additional epoch of contrastive fine-tuning to better align with the evaluation process. Based on prior studies and empirical observations, the quality of negative samples plays a crucial role in training effectiveness. To enhance negative sampling, we adopt NV-Embed for hard negative mining. Specifically, we randomly select 7 hard negatives from the ranked list positions 30 to 210, treating these as challenging negative samples during training.

\paragraph{Baselines} We categorize the baselines into three groups: (1) models without contrast training, including base models with various embedding methods using the same instructions as AutoRegEmbed and prompt-adjusted embedded models, including Echo \citep{DBLP:journals/corr/abs-2402-15449}, PromptEOL \citep{DBLP:conf/emnlp/JiangHLWZ24}, MetaEOL \citep{DBLP:conf/acl/LeiW00CTY24}, and GenEOL \citep{DBLP:journals/corr/abs-2410-14635}; (2) unsupervised contrast training models, primarily LLM2Vec \citep{DBLP:journals/corr/abs-2404-05961} with different base models; and (3) supervised contrast training models, which consist of NV-Embed \citep{DBLP:journals/corr/abs-2405-17428}, SFR-Embedding-2\_R \citep{SFR-embedding-2}, gte-Qwen2-7B-instruct \citep{DBLP:journals/corr/abs-2308-03281}, LLM2Vec \citep{DBLP:journals/corr/abs-2404-05961}, and fair baselines.

\subsection{Main Results}
Table~\ref{tab:sts} summarizes the results of various baselines and AutoRegEmbed on ten STS datasets, along with the training data required for each method.

\paragraph{AutoRegEmbed vs. Without Contrastive Training} Models without contrastive training are divided into two categories. The first is our own fair baseline model, which performs significantly worse than AutoRegEmbed, with an average performance 20\% lower. This highlights the difficulty of untrained LLMs in directly generating high-quality embeddings. While some methods enhance the base model’s embeddings through prompt optimization, their improvements remain limited—even on a 13B-parameter model—and come with significant additional reasoning costs.

\paragraph{AutoRegEmbed vs. Unsupervised Contrastive Training}
LLM2Vec enhances existing LLMs using an unsupervised contrastive learning approach similar to SimCSE, leading to significant performance gains. Compared to the base model, the unsupervised version of LLM2Vec improves performance by over 15\%. Although it utilizes almost 160,000 data samples, its performance remains 4.74\% lower than AutoRegEmbed, demonstrating its lower efficiency.

\paragraph{AutoRegEmbed vs. Supervised Contrastive Training}
Supervised contrastive learning is the mainstream approach for building high-quality embedding models. We first compared SOTA methods that once ranked on the MTEB leaderbord, including NV-Embed, SFR-Embedding-2\_R, gte-Qwen2-7B-instruct, and LLM2Vec. In terms of performance, AutoRegEmbed surpasses all leading methods across 10 STS datasets, outperforming LLM2Vec by a margin of 0.58. From a data efficiency perspective, AutoRegEmbed achieves performance comparable to the previous SOTA models with just 66,382 training samples, whereas the latter requires tens of millions of triplets to reach peak performance. Additionally, previous SOTA models employ multi-task learning (e.g., retrieval and clustering), whose impact on STS performance remains unclear. To ensure a fair comparison, we use single-task contrastive learning as a baseline. Unlike traditional contrastive learning, AutoRegEmbed does not rely on in-batch negative samples. So we add two baselines to single-task contrastive learning that also exclude the in-batch negative sample strategy. As shown in Table~\ref{tab:sts}, even under identical training data, AutoRegEmbed outperforms four different single-task contrastive learning, further validating its effectiveness.

\subsection{Performance on retrieval tasks}
Table~\ref{tab:retrieval_datasets} summarizes the performance of various baseline methods and AutoRegEmbed across three retrieval datasets. On MS MARCO, AutoRegEmbed outperforms most prior state-of-the-art (SOTA) models and ranks second only to gte-Qwen2-7B-instruct. It also achieves competitive results on the other two datasets, NFcorpus and SCIDOCS. This outcome is expected, as AutoRegEmbed is trained solely on MS MARCO, whereas previous SOTA models are trained on large-scale datasets spanning diverse distributions, which contributes to their strong generalization on NFcorpus and SCIDOCS.

Notably, AutoRegEmbed consistently outperforms LLaMA2-inbatch-M, which is trained on the same data, across all three datasets. This further confirms the effectiveness of our method in retrieval tasks, even without large-scale multi-domain training.
\begin{table}[ht]
\centering
\small
\setlength{\tabcolsep}{8pt}
\renewcommand{\arraystretch}{1.3}
\resizebox{\linewidth}{!}{%
\begin{tabular}{lccc}
\toprule
\textbf{Method} & \textbf{MS MARCO} & \textbf{NFcorpus} & \textbf{SCIDOCS} \\
\midrule
LLM2Vec(Unsupervised) & 18.81 & 26.81 & 10.00 \\
LLM2Vec(Supervised) & 41.45 & 40.33 &  21.05\\
SFR-Embedding-2\_R & 42.18 & \textbf{41.34}  & \textbf{24.69} \\
gte-Qwen2-7B-instruct & \textbf{45.98} & \underline{40.60} & \underline{23.48} \\
LLaMA2-inbatch-M & 41.67 & 34.19 & 16.15 \\
AutoRegEmbed & \underline{42.49} & 38.16 & 19.79 \\
\bottomrule
\end{tabular}
}
\caption{Evaluation results on MS MARCO, NFcorpus, and SCIDOCS.}
\label{tab:retrieval_datasets}
\end{table}

\subsection{Ablation Study}
To verify the effectiveness of AutoRegEmbed, we conducted an ablation study. First, we removed Conditional Distribution Alignment to evaluate its impact on model performance. Second, since Equation~\ref{eq7} was derived empirically in our previous work, we tested different variants of this equation to confirm that it remains the optimal choice. Different variants of Equation~\ref{eq7} include \textbf{Log\_sigmoid}, which maps similarity to a logarithmic scale for integration with the exponential function e, as well as \textbf{KL divergence} and \textbf{JS divergence}, which quantify the distance between the conditional probability distributions of positive and negative sample embeddings in distinct ways. 

\begin{table}
\centering
\normalsize
\resizebox{\linewidth}{!}{%
\begin{tabular}{lc}
\toprule
\textbf{Method} & \textbf{Avg.}\\
\midrule
{AutoRegEmbed-LLaMA2} &83.24$_{(10)}$\\
\midrule
\multicolumn{2}{l}{\it{Tasks}}\\
w/o Conditional Distribution Alignment & 73.90$_{(10)}$\\
LLaMA2-L (Without Training) & 56.91$_{(10)}$\\
\midrule
\multicolumn{2}{l}{\it{Equation~\ref{eq7}}}\\
Log\_sigmoid & 82.93$_{(10)}$\\
KL divergence & 79.82$_{(10)}$\\
JS divergence & 79.02$_{(10)}$ \\
\bottomrule
\end{tabular}
}
\caption{Ablation experiments of AutoRegEmbed. We conduct ablation and contrast experiments on various tasks and Equation~\ref{eq7} to demonstrate the effectiveness of AutoRegEmbed.}
\label{tab:ablation_study}
\end{table}

\begin{figure}[ht]
  \includegraphics[width=0.48\textwidth]{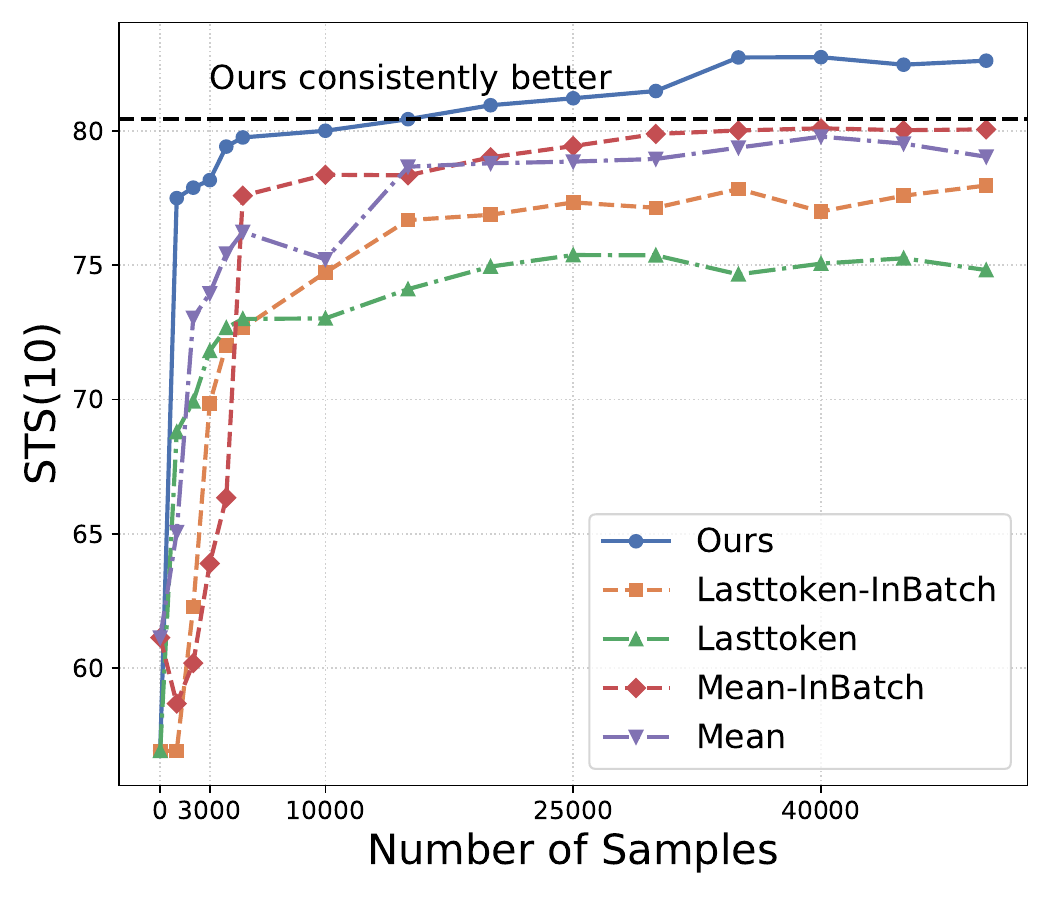}
  \caption{We evaluate the learning efficiency of our method against traditional contrastive learning on 10 STS datasets, comparing their performance under the same number of samples. Further details are provided in Appendix~\ref{app:details}.}
  \label{fig:convergence}
  \vspace{-5pt}%
\end{figure}

Table~\ref{tab:ablation_study} presents the ablation results. The experiments on different tasks indicate that Conditional Distribution Alignment improves performance by 9.17\%, while Information Compression contributes a 16.99\% improvement, demonstrating the effectiveness of both tasks. Additionally, experiments on variants of Equation~\ref{eq7} reveal that, although using a logarithmic scale for similarity and employing KL or JS divergence to measure distribution distance are more intuitive approaches, they do not surpass the performance of the original loss function in Equation~\ref{eq7}. Thus, Equation~\ref{eq7} can be regarded as a more effective loss function. The specific equations and more detailed analysis are provided in Appendix~\ref{app:ablation}.

\subsection{Learning Efficiency}

To verify that AutoRegEmbed is better suited for LLMs, we compare its performance with four contrastive learning baselines under the same training data. Figure~\ref{fig:convergence} shows that as the training data increases, the performance of both AutoRegEmbed and other contrastive learning methods improves, but AutoRegEmbed exhibits the fastest growth. Notably, with just 15,000 samples, AutoRegEmbed already surpasses the maximum performance of other contrastive learning models.  The results indicate that AutoRegEmbed significantly outperforms the baseline models in learning efficiency.

\section{Conclusions}
To address the limitation that traditional contrastive learning does not adhere to the autoregressive nature of LLMs, we propose AutoRegEmbed—a novel contrastive learning method based on embedded conditional probability distributions. AutoRegEmbed ensures that LLM-generated embeddings capture global semantics while maintaining alignment and uniformity through information compression and conditional distribution alignment tasks. AutoRegEmbed achieves comparable performance to SOTA models with fewer training samples and superior learning efficiency.
\section{Limitations}
The primary advantage of AutoRegEmbed lies in its ability to effectively harness the power of large language models (LLMs) to construct robust and high-quality text embeddings. However, it is important to acknowledge several limitations of our approach.

AutoRegEmbed does not possess inherent mechanisms to filter or detect malicious or harmful content in the data it processes. While the model is capable of generating embeddings from a wide range of text inputs, it lacks the ability to evaluate the ethical or safety implications of the data. This makes it vulnerable to issues related to biased, offensive, or otherwise problematic content present in the training corpus. In cases where the training data contains harmful or discriminatory material, the embeddings generated by AutoRegEmbed may inadvertently carry forward these biases, potentially leading to unintended and undesirable outcomes when applied to real-world tasks.

To mitigate this risk, we recommend that users of AutoRegEmbed ensure that the training data is carefully curated, and ideally, filtered for harmful content. Additionally, users should be cautious when applying AutoRegEmbed to sensitive domains, where the generation of unsafe or biased embeddings could have significant consequences.
\section*{Acknowledgments}
This work was supported by the Strategic Priority Research Program of the CAS under Grants No.XDB0680302, the National Natural Science Foundation of China (NSFC) under Grants No. 62276248, the Key Research and Development Program of Xinjiang Uyghur Autonomous Region Grant No. 2024B03026, the Beijing Nova Program under Grants No. 20250484765, and the Youth Innovation Promotion Association CAS under Grants No. 2023111.

\bibliography{acl_latex}

\begin{thebibliography}{64}
\providecommand{\natexlab}[1]{#1}

\bibitem[{Agirre et~al.(2015)Agirre, Banea, Cardie, Cer, Diab, Gonzalez{-}Agirre, Guo, Lopez{-}Gazpio, Maritxalar, Mihalcea, Rigau, Uria, and Wiebe}]{DBLP:conf/semeval/AgirreBCCDGGLMM15}
Eneko Agirre, Carmen Banea, Claire Cardie, Daniel~M. Cer, Mona~T. Diab, Aitor Gonzalez{-}Agirre, Weiwei Guo, I{\~{n}}igo Lopez{-}Gazpio, Montse Maritxalar, Rada Mihalcea, German Rigau, Larraitz Uria, and Janyce Wiebe. 2015.
\newblock \href {https://doi.org/10.18653/V1/S15-2045} {Semeval-2015 task 2: Semantic textual similarity, english, spanish and pilot on interpretability}.
\newblock In \emph{Proceedings of the 9th International Workshop on Semantic Evaluation, SemEval@NAACL-HLT 2015, Denver, Colorado, USA, June 4-5, 2015}, pages 252--263. The Association for Computer Linguistics.

\bibitem[{Agirre et~al.(2014)Agirre, Banea, Cardie, Cer, Diab, Gonzalez{-}Agirre, Guo, Mihalcea, Rigau, and Wiebe}]{DBLP:conf/semeval/AgirreBCCDGGMRW14}
Eneko Agirre, Carmen Banea, Claire Cardie, Daniel~M. Cer, Mona~T. Diab, Aitor Gonzalez{-}Agirre, Weiwei Guo, Rada Mihalcea, German Rigau, and Janyce Wiebe. 2014.
\newblock \href {https://doi.org/10.3115/V1/S14-2010} {Semeval-2014 task 10: Multilingual semantic textual similarity}.
\newblock In \emph{Proceedings of the 8th International Workshop on Semantic Evaluation, SemEval@COLING 2014, Dublin, Ireland, August 23-24, 2014}, pages 81--91. The Association for Computer Linguistics.

\bibitem[{Agirre et~al.(2016)Agirre, Banea, Cer, Diab, Gonzalez{-}Agirre, Mihalcea, Rigau, and Wiebe}]{DBLP:conf/semeval/AgirreBCDGMRW16}
Eneko Agirre, Carmen Banea, Daniel~M. Cer, Mona~T. Diab, Aitor Gonzalez{-}Agirre, Rada Mihalcea, German Rigau, and Janyce Wiebe. 2016.
\newblock \href {https://doi.org/10.18653/V1/S16-1081} {Semeval-2016 task 1: Semantic textual similarity, monolingual and cross-lingual evaluation}.
\newblock In \emph{Proceedings of the 10th International Workshop on Semantic Evaluation, SemEval@NAACL-HLT 2016, San Diego, CA, USA, June 16-17, 2016}, pages 497--511. The Association for Computer Linguistics.

\bibitem[{Agirre et~al.(2012)Agirre, Cer, Diab, and Gonzalez{-}Agirre}]{DBLP:conf/semeval/AgirreCDG12}
Eneko Agirre, Daniel~M. Cer, Mona~T. Diab, and Aitor Gonzalez{-}Agirre. 2012.
\newblock \href {https://aclanthology.org/S12-1051/} {Semeval-2012 task 6: {A} pilot on semantic textual similarity}.
\newblock In \emph{Proceedings of the 6th International Workshop on Semantic Evaluation, SemEval@NAACL-HLT 2012, Montr{\'{e}}al, Canada, June 7-8, 2012}, pages 385--393. The Association for Computer Linguistics.

\bibitem[{Agirre et~al.(2013)Agirre, Cer, Diab, Gonzalez{-}Agirre, and Guo}]{DBLP:conf/starsem/AgirreCDGG13}
Eneko Agirre, Daniel~M. Cer, Mona~T. Diab, Aitor Gonzalez{-}Agirre, and Weiwei Guo. 2013.
\newblock \href {https://aclanthology.org/S13-1004/} {*sem 2013 shared task: Semantic textual similarity}.
\newblock In \emph{Proceedings of the Second Joint Conference on Lexical and Computational Semantics, *SEM 2013, June 13-14, 2013, Atlanta, Georgia, {USA}}, pages 32--43. Association for Computational Linguistics.

\bibitem[{Asai et~al.(2023)Asai, Schick, Lewis, Chen, Izacard, Riedel, Hajishirzi, and Yih}]{DBLP:conf/acl/AsaiSL0I0HY23}
Akari Asai, Timo Schick, Patrick S.~H. Lewis, Xilun Chen, Gautier Izacard, Sebastian Riedel, Hannaneh Hajishirzi, and Wen{-}tau Yih. 2023.
\newblock \href {https://doi.org/10.18653/V1/2023.FINDINGS-ACL.225} {Task-aware retrieval with instructions}.
\newblock In \emph{Findings of the Association for Computational Linguistics: {ACL} 2023, Toronto, Canada, July 9-14, 2023}, pages 3650--3675. Association for Computational Linguistics.

\bibitem[{BehnamGhader et~al.(2024)BehnamGhader, Adlakha, Mosbach, Bahdanau, Chapados, and Reddy}]{DBLP:journals/corr/abs-2404-05961}
Parishad BehnamGhader, Vaibhav Adlakha, Marius Mosbach, Dzmitry Bahdanau, Nicolas Chapados, and Siva Reddy. 2024.
\newblock \href {https://doi.org/10.48550/ARXIV.2404.05961} {Llm2vec: Large language models are secretly powerful text encoders}.
\newblock \emph{CoRR}, abs/2404.05961.

\bibitem[{Beigi et~al.(2024)Beigi, Tan, Mudiam, Chen, Shu, and Liu}]{DBLP:conf/dsaa/BeigiTMCS024}
Alimohammad Beigi, Zhen Tan, Nivedh Mudiam, Canyu Chen, Kai Shu, and Huan Liu. 2024.
\newblock \href {https://doi.org/10.1109/DSAA61799.2024.10722818} {Model attribution in llm-generated disinformation: {A} domain generalization approach with supervised contrastive learning}.
\newblock In \emph{11th {IEEE} International Conference on Data Science and Advanced Analytics, {DSAA} 2024, San Diego, CA, USA, October 6-10, 2024}, pages 1--10. {IEEE}.

\bibitem[{Boteva et~al.(2016)Boteva, Ghalandari, Sokolov, and Riezler}]{DBLP:conf/ecir/BotevaGSR16}
Vera Boteva, Demian~Gholipour Ghalandari, Artem Sokolov, and Stefan Riezler. 2016.
\newblock \href {https://doi.org/10.1007/978-3-319-30671-1\_58} {A full-text learning to rank dataset for medical information retrieval}.
\newblock In \emph{Advances in Information Retrieval - 38th European Conference on {IR} Research, {ECIR} 2016, Padua, Italy, March 20-23, 2016. Proceedings}, volume 9626 of \emph{Lecture Notes in Computer Science}, pages 716--722. Springer.

\bibitem[{Cer et~al.(2017)Cer, Diab, Agirre, Lopez{-}Gazpio, and Specia}]{DBLP:conf/semeval/CerDALS17}
Daniel~M. Cer, Mona~T. Diab, Eneko Agirre, I{\~{n}}igo Lopez{-}Gazpio, and Lucia Specia. 2017.
\newblock \href {https://doi.org/10.18653/V1/S17-2001} {Semeval-2017 task 1: Semantic textual similarity multilingual and crosslingual focused evaluation}.
\newblock In \emph{Proceedings of the 11th International Workshop on Semantic Evaluation, SemEval@ACL 2017, Vancouver, Canada, August 3-4, 2017}, pages 1--14. Association for Computational Linguistics.

\bibitem[{Chen et~al.(2024{\natexlab{a}})Chen, Dong, Shu, Zhang, Sesay, Karlsson, Fu, and Shi}]{DBLP:conf/ijcai/ChenDSZS00024}
Guangyao Chen, Siwei Dong, Yu~Shu, Ge~Zhang, Jaward Sesay, B{\"{o}}rje Karlsson, Jie Fu, and Yemin Shi. 2024{\natexlab{a}}.
\newblock \href {https://www.ijcai.org/proceedings/2024/3} {Autoagents: {A} framework for automatic agent generation}.
\newblock In \emph{Proceedings of the Thirty-Third International Joint Conference on Artificial Intelligence, {IJCAI} 2024, Jeju, South Korea, August 3-9, 2024}, pages 22--30. ijcai.org.

\bibitem[{Chen et~al.(2024{\natexlab{b}})Chen, Xiao, Zhang, Luo, Lian, and Liu}]{DBLP:journals/corr/abs-2402-03216}
Jianlv Chen, Shitao Xiao, Peitian Zhang, Kun Luo, Defu Lian, and Zheng Liu. 2024{\natexlab{b}}.
\newblock \href {https://doi.org/10.48550/ARXIV.2402.03216} {{BGE} m3-embedding: Multi-lingual, multi-functionality, multi-granularity text embeddings through self-knowledge distillation}.
\newblock \emph{CoRR}, abs/2402.03216.

\bibitem[{Chen et~al.(2022)Chen, Zeynali, Camargo, Fl{\"{o}}ck, Gaffney, Grabowicz, Hale, Jurgens, and Samory}]{DBLP:conf/semeval/ChenZCFGGHJS22}
Xi~Chen, Ali Zeynali, Chico~Q. Camargo, Fabian Fl{\"{o}}ck, Devin Gaffney, Przemyslaw~A. Grabowicz, Scott Hale, David Jurgens, and Mattia Samory. 2022.
\newblock \href {https://doi.org/10.18653/V1/2022.SEMEVAL-1.155} {Semeval-2022 task 8: Multilingual news article similarity}.
\newblock In \emph{Proceedings of the 16th International Workshop on Semantic Evaluation, SemEval@NAACL 2022, Seattle, Washington, United States, July 14-15, 2022}, pages 1094--1106. Association for Computational Linguistics.

\bibitem[{Chevalier et~al.(2023)Chevalier, Wettig, Ajith, and Chen}]{DBLP:conf/emnlp/ChevalierWAC23}
Alexis Chevalier, Alexander Wettig, Anirudh Ajith, and Danqi Chen. 2023.
\newblock \href {https://doi.org/10.18653/V1/2023.EMNLP-MAIN.232} {Adapting language models to compress contexts}.
\newblock In \emph{Proceedings of the 2023 Conference on Empirical Methods in Natural Language Processing, {EMNLP} 2023, Singapore, December 6-10, 2023}, pages 3829--3846. Association for Computational Linguistics.

\bibitem[{Cohan et~al.(2020)Cohan, Feldman, Beltagy, Downey, and Weld}]{DBLP:conf/acl/CohanFBDW20}
Arman Cohan, Sergey Feldman, Iz~Beltagy, Doug Downey, and Daniel~S. Weld. 2020.
\newblock \href {https://doi.org/10.18653/V1/2020.ACL-MAIN.207} {{SPECTER:} document-level representation learning using citation-informed transformers}.
\newblock In \emph{Proceedings of the 58th Annual Meeting of the Association for Computational Linguistics, {ACL} 2020, Online, July 5-10, 2020}, pages 2270--2282. Association for Computational Linguistics.

\bibitem[{Dao(2024)}]{DBLP:conf/iclr/Dao24}
Tri Dao. 2024.
\newblock \href {https://openreview.net/forum?id=mZn2Xyh9Ec} {Flashattention-2: Faster attention with better parallelism and work partitioning}.
\newblock In \emph{The Twelfth International Conference on Learning Representations, {ICLR} 2024, Vienna, Austria, May 7-11, 2024}. OpenReview.net.

\bibitem[{Deng et~al.(2022)Deng, Dai, Guo, Ju, and Peng}]{DBLP:conf/emnlp/DengDGJP22}
Jingcheng Deng, Hengwei Dai, Xuewei Guo, Yuanchen Ju, and Wei Peng. 2022.
\newblock \href {https://doi.org/10.18653/V1/2022.EMNLP-MAIN.584} {{IRRGN:} an implicit relational reasoning graph network for multi-turn response selection}.
\newblock In \emph{Proceedings of the 2022 Conference on Empirical Methods in Natural Language Processing, {EMNLP} 2022, Abu Dhabi, United Arab Emirates, December 7-11, 2022}, pages 8529--8541. Association for Computational Linguistics.

\bibitem[{Deng et~al.(2023)Deng, Pang, Shen, and Cheng}]{DBLP:conf/emnlp/DengPSC23}
Jingcheng Deng, Liang Pang, Huawei Shen, and Xueqi Cheng. 2023.
\newblock \href {https://doi.org/10.18653/V1/2023.FINDINGS-EMNLP.164} {Regavae: {A} retrieval-augmented gaussian mixture variational auto-encoder for language modeling}.
\newblock In \emph{Findings of the Association for Computational Linguistics: {EMNLP} 2023, Singapore, December 6-10, 2023}, pages 2500--2510. Association for Computational Linguistics.

\bibitem[{Devlin et~al.(2019)Devlin, Chang, Lee, and Toutanova}]{DBLP:conf/naacl/DevlinCLT19}
Jacob Devlin, Ming{-}Wei Chang, Kenton Lee, and Kristina Toutanova. 2019.
\newblock \href {https://doi.org/10.18653/V1/N19-1423} {{BERT:} pre-training of deep bidirectional transformers for language understanding}.
\newblock In \emph{Proceedings of the 2019 Conference of the North American Chapter of the Association for Computational Linguistics: Human Language Technologies, {NAACL-HLT} 2019, Minneapolis, MN, USA, June 2-7, 2019, Volume 1 (Long and Short Papers)}, pages 4171--4186. Association for Computational Linguistics.

\bibitem[{Ding et~al.(2024)Ding, Pang, Wei, Shen, and Cheng}]{DBLP:journals/corr/abs-2402-10612}
Hanxing Ding, Liang Pang, Zihao Wei, Huawei Shen, and Xueqi Cheng. 2024.
\newblock \href {https://doi.org/10.48550/ARXIV.2402.10612} {Retrieve only when it needs: Adaptive retrieval augmentation for hallucination mitigation in large language models}.
\newblock \emph{CoRR}, abs/2402.10612.

\bibitem[{Duan et~al.(2025)Duan, Duan, Yin, Shen, Jing, Zhang, Shen, and Cheng}]{duan-etal-2025-related}
Zenghao Duan, Wenbin Duan, Zhiyi Yin, Yinghan Shen, Shaoling Jing, Jie Zhang, Huawei Shen, and Xueqi Cheng. 2025.
\newblock \href {https://aclanthology.org/2025.naacl-short.31/} {Related knowledge perturbation matters: Rethinking multiple pieces of knowledge editing in same-subject}.
\newblock In \emph{Proceedings of the 2025 Conference of the Nations of the Americas Chapter of the Association for Computational Linguistics: Human Language Technologies (Volume 2: Short Papers)}, pages 363--373, Albuquerque, New Mexico. Association for Computational Linguistics.

\bibitem[{Gao and Callan(2022)}]{DBLP:conf/acl/GaoC22}
Luyu Gao and Jamie Callan. 2022.
\newblock \href {https://doi.org/10.18653/V1/2022.ACL-LONG.203} {Unsupervised corpus aware language model pre-training for dense passage retrieval}.
\newblock In \emph{Proceedings of the 60th Annual Meeting of the Association for Computational Linguistics (Volume 1: Long Papers), {ACL} 2022, Dublin, Ireland, May 22-27, 2022}, pages 2843--2853. Association for Computational Linguistics.

\bibitem[{Gao et~al.(2021)Gao, Yao, and Chen}]{DBLP:conf/emnlp/GaoYC21}
Tianyu Gao, Xingcheng Yao, and Danqi Chen. 2021.
\newblock \href {https://doi.org/10.18653/V1/2021.EMNLP-MAIN.552} {Simcse: Simple contrastive learning of sentence embeddings}.
\newblock In \emph{Proceedings of the 2021 Conference on Empirical Methods in Natural Language Processing, {EMNLP} 2021, Virtual Event / Punta Cana, Dominican Republic, 7-11 November, 2021}, pages 6894--6910. Association for Computational Linguistics.

\bibitem[{Ge et~al.(2024)Ge, Hu, Wang, Wang, Chen, and Wei}]{DBLP:conf/iclr/00010WWCW24}
Tao Ge, Jing Hu, Lei Wang, Xun Wang, Si{-}Qing Chen, and Furu Wei. 2024.
\newblock \href {https://openreview.net/forum?id=uREj4ZuGJE} {In-context autoencoder for context compression in a large language model}.
\newblock In \emph{The Twelfth International Conference on Learning Representations, {ICLR} 2024, Vienna, Austria, May 7-11, 2024}. OpenReview.net.

\bibitem[{He et~al.(2021)He, Liu, Gao, and Chen}]{DBLP:conf/iclr/HeLGC21}
Pengcheng He, Xiaodong Liu, Jianfeng Gao, and Weizhu Chen. 2021.
\newblock \href {https://openreview.net/forum?id=XPZIaotutsD} {Deberta: decoding-enhanced bert with disentangled attention}.
\newblock In \emph{9th International Conference on Learning Representations, {ICLR} 2021, Virtual Event, Austria, May 3-7, 2021}. OpenReview.net.

\bibitem[{Hu et~al.(2022)Hu, Shen, Wallis, Allen{-}Zhu, Li, Wang, Wang, and Chen}]{DBLP:conf/iclr/HuSWALWWC22}
Edward~J. Hu, Yelong Shen, Phillip Wallis, Zeyuan Allen{-}Zhu, Yuanzhi Li, Shean Wang, Lu~Wang, and Weizhu Chen. 2022.
\newblock \href {https://openreview.net/forum?id=nZeVKeeFYf9} {Lora: Low-rank adaptation of large language models}.
\newblock In \emph{The Tenth International Conference on Learning Representations, {ICLR} 2022, Virtual Event, April 25-29, 2022}. OpenReview.net.

\bibitem[{Jiang et~al.(2024)Jiang, Huang, Luan, Wang, and Zhuang}]{DBLP:conf/emnlp/JiangHLWZ24}
Ting Jiang, Shaohan Huang, Zhongzhi Luan, Deqing Wang, and Fuzhen Zhuang. 2024.
\newblock \href {https://aclanthology.org/2024.findings-emnlp.181} {Scaling sentence embeddings with large language models}.
\newblock In \emph{Findings of the Association for Computational Linguistics: {EMNLP} 2024, Miami, Florida, USA, November 12-16, 2024}, pages 3182--3196. Association for Computational Linguistics.

\bibitem[{Karpukhin et~al.(2020)Karpukhin, Oguz, Min, Lewis, Wu, Edunov, Chen, and Yih}]{DBLP:conf/emnlp/KarpukhinOMLWEC20}
Vladimir Karpukhin, Barlas Oguz, Sewon Min, Patrick S.~H. Lewis, Ledell Wu, Sergey Edunov, Danqi Chen, and Wen{-}tau Yih. 2020.
\newblock \href {https://doi.org/10.18653/V1/2020.EMNLP-MAIN.550} {Dense passage retrieval for open-domain question answering}.
\newblock In \emph{Proceedings of the 2020 Conference on Empirical Methods in Natural Language Processing, {EMNLP} 2020, Online, November 16-20, 2020}, pages 6769--6781. Association for Computational Linguistics.

\bibitem[{Khandelwal et~al.(2020)Khandelwal, Levy, Jurafsky, Zettlemoyer, and Lewis}]{DBLP:conf/iclr/KhandelwalLJZL20}
Urvashi Khandelwal, Omer Levy, Dan Jurafsky, Luke Zettlemoyer, and Mike Lewis. 2020.
\newblock \href {https://openreview.net/forum?id=HklBjCEKvH} {Generalization through memorization: Nearest neighbor language models}.
\newblock In \emph{8th International Conference on Learning Representations, {ICLR} 2020, Addis Ababa, Ethiopia, April 26-30, 2020}. OpenReview.net.

\bibitem[{Lee et~al.(2024)Lee, Roy, Xu, Raiman, Shoeybi, Catanzaro, and Ping}]{DBLP:journals/corr/abs-2405-17428}
Chankyu Lee, Rajarshi Roy, Mengyao Xu, Jonathan Raiman, Mohammad Shoeybi, Bryan Catanzaro, and Wei Ping. 2024.
\newblock \href {https://doi.org/10.48550/ARXIV.2405.17428} {Nv-embed: Improved techniques for training llms as generalist embedding models}.
\newblock \emph{CoRR}, abs/2405.17428.

\bibitem[{Lei et~al.(2024)Lei, Wu, Zhou, Shen, Cao, Tao, and Yates}]{DBLP:conf/acl/LeiW00CTY24}
Yibin Lei, Di~Wu, Tianyi Zhou, Tao Shen, Yu~Cao, Chongyang Tao, and Andrew Yates. 2024.
\newblock \href {https://doi.org/10.18653/V1/2024.ACL-LONG.546} {Meta-task prompting elicits embeddings from large language models}.
\newblock In \emph{Proceedings of the 62nd Annual Meeting of the Association for Computational Linguistics (Volume 1: Long Papers), {ACL} 2024, Bangkok, Thailand, August 11-16, 2024}, pages 10141--10157. Association for Computational Linguistics.

\bibitem[{Li et~al.(2020)Li, Zhou, He, Wang, Yang, and Li}]{DBLP:conf/emnlp/LiZHWYL20}
Bohan Li, Hao Zhou, Junxian He, Mingxuan Wang, Yiming Yang, and Lei Li. 2020.
\newblock \href {https://doi.org/10.18653/V1/2020.EMNLP-MAIN.733} {On the sentence embeddings from pre-trained language models}.
\newblock In \emph{Proceedings of the 2020 Conference on Empirical Methods in Natural Language Processing, {EMNLP} 2020, Online, November 16-20, 2020}, pages 9119--9130. Association for Computational Linguistics.

\bibitem[{Li et~al.(2024{\natexlab{a}})Li, Liu, Xiao, Shao, and Lian}]{DBLP:conf/acl/Li0XSL24}
Chaofan Li, Zheng Liu, Shitao Xiao, Yingxia Shao, and Defu Lian. 2024{\natexlab{a}}.
\newblock \href {https://doi.org/10.18653/V1/2024.ACL-LONG.191} {Llama2vec: Unsupervised adaptation of large language models for dense retrieval}.
\newblock In \emph{Proceedings of the 62nd Annual Meeting of the Association for Computational Linguistics (Volume 1: Long Papers), {ACL} 2024, Bangkok, Thailand, August 11-16, 2024}, pages 3490--3500. Association for Computational Linguistics.

\bibitem[{Li et~al.(2024{\natexlab{b}})Li, Qin, Xiao, Chen, Luo, Shao, Lian, and Liu}]{DBLP:journals/corr/abs-2409-15700}
Chaofan Li, Minghao Qin, Shitao Xiao, Jianlyu Chen, Kun Luo, Yingxia Shao, Defu Lian, and Zheng Liu. 2024{\natexlab{b}}.
\newblock \href {https://doi.org/10.48550/ARXIV.2409.15700} {Making text embedders few-shot learners}.
\newblock \emph{CoRR}, abs/2409.15700.

\bibitem[{Li et~al.(2023)Li, Zhang, Zhang, Long, Xie, and Zhang}]{DBLP:journals/corr/abs-2308-03281}
Zehan Li, Xin Zhang, Yanzhao Zhang, Dingkun Long, Pengjun Xie, and Meishan Zhang. 2023.
\newblock \href {https://doi.org/10.48550/ARXIV.2308.03281} {Towards general text embeddings with multi-stage contrastive learning}.
\newblock \emph{CoRR}, abs/2308.03281.

\bibitem[{Liu et~al.(2025)Liu, Sheng, Wang, Li, Yang, and Cao}]{liu2025forewarnedforearmedpresynthesizingjailbreaklike}
Sheng Liu, Qiang Sheng, Danding Wang, Yang Li, Guang Yang, and Juan Cao. 2025.
\newblock \href {https://arxiv.org/abs/2508.20038} {Forewarned is forearmed: Pre-synthesizing jailbreak-like instructions to enhance llm safety guardrail to potential attacks}.
\newblock \emph{Preprint}, arXiv:2508.20038.

\bibitem[{Liu et~al.(2024)Liu, Trager, Achille, Perera, Zancato, and Soatto}]{DBLP:conf/iclr/LiuTAPZS24}
Tian~Yu Liu, Matthew Trager, Alessandro Achille, Pramuditha Perera, Luca Zancato, and Stefano Soatto. 2024.
\newblock \href {https://openreview.net/forum?id=UyGWafcopT} {Meaning representations from trajectories in autoregressive models}.
\newblock In \emph{The Twelfth International Conference on Learning Representations, {ICLR} 2024, Vienna, Austria, May 7-11, 2024}. OpenReview.net.

\bibitem[{Ma et~al.(2024)Ma, Wang, Yang, Wei, and Lin}]{DBLP:conf/sigir/MaWYWL24}
Xueguang Ma, Liang Wang, Nan Yang, Furu Wei, and Jimmy Lin. 2024.
\newblock \href {https://doi.org/10.1145/3626772.3657951} {Fine-tuning llama for multi-stage text retrieval}.
\newblock In \emph{Proceedings of the 47th International {ACM} {SIGIR} Conference on Research and Development in Information Retrieval, {SIGIR} 2024, Washington DC, USA, July 14-18, 2024}, pages 2421--2425. {ACM}.

\bibitem[{Meng et~al.(2024)Meng, Liu, Joty, Xiong, Zhou, and Yavuz}]{SFR-embedding-2}
Rui Meng, Ye~Liu, Shafiq~Rayhan Joty, Caiming Xiong, Yingbo Zhou, and Semih Yavuz. 2024.
\newblock \href {https://huggingface.co/Salesforce/SFR-Embedding-2_R} {Sfr-embedding-2: Advanced text embedding with multi-stage training}.

\bibitem[{Mu et~al.(2023)Mu, Li, and Goodman}]{DBLP:conf/nips/Mu0G23}
Jesse Mu, Xiang Li, and Noah~D. Goodman. 2023.
\newblock \href {http://papers.nips.cc/paper\_files/paper/2023/hash/3d77c6dcc7f143aa2154e7f4d5e22d68-Abstract-Conference.html} {Learning to compress prompts with gist tokens}.
\newblock In \emph{Advances in Neural Information Processing Systems 36: Annual Conference on Neural Information Processing Systems 2023, NeurIPS 2023, New Orleans, LA, USA, December 10 - 16, 2023}.

\bibitem[{Muennighoff et~al.(2024)Muennighoff, Su, Wang, Yang, Wei, Yu, Singh, and Kiela}]{DBLP:journals/corr/abs-2402-09906}
Niklas Muennighoff, Hongjin Su, Liang Wang, Nan Yang, Furu Wei, Tao Yu, Amanpreet Singh, and Douwe Kiela. 2024.
\newblock \href {https://doi.org/10.48550/ARXIV.2402.09906} {Generative representational instruction tuning}.
\newblock \emph{CoRR}, abs/2402.09906.

\bibitem[{Muennighoff et~al.(2023)Muennighoff, Tazi, Magne, and Reimers}]{DBLP:conf/eacl/MuennighoffTMR23}
Niklas Muennighoff, Nouamane Tazi, Lo{\"{\i}}c Magne, and Nils Reimers. 2023.
\newblock \href {https://doi.org/10.18653/V1/2023.EACL-MAIN.148} {{MTEB:} massive text embedding benchmark}.
\newblock In \emph{Proceedings of the 17th Conference of the European Chapter of the Association for Computational Linguistics, {EACL} 2023, Dubrovnik, Croatia, May 2-6, 2023}, pages 2006--2029. Association for Computational Linguistics.

\bibitem[{Nguyen et~al.(2016)Nguyen, Rosenberg, Song, Gao, Tiwary, Majumder, and Deng}]{DBLP:conf/nips/NguyenRSGTMD16}
Tri Nguyen, Mir Rosenberg, Xia Song, Jianfeng Gao, Saurabh Tiwary, Rangan Majumder, and Li~Deng. 2016.
\newblock \href {https://ceur-ws.org/Vol-1773/CoCoNIPS\_2016\_paper9.pdf} {{MS} {MARCO:} {A} human generated machine reading comprehension dataset}.
\newblock In \emph{Proceedings of the Workshop on Cognitive Computation: Integrating neural and symbolic approaches 2016 co-located with the 30th Annual Conference on Neural Information Processing Systems {(NIPS} 2016), Barcelona, Spain, December 9, 2016}, volume 1773 of \emph{{CEUR} Workshop Proceedings}. CEUR-WS.org.

\bibitem[{Rafailov et~al.(2023)Rafailov, Sharma, Mitchell, Manning, Ermon, and Finn}]{DBLP:conf/nips/RafailovSMMEF23}
Rafael Rafailov, Archit Sharma, Eric Mitchell, Christopher~D. Manning, Stefano Ermon, and Chelsea Finn. 2023.
\newblock \href {http://papers.nips.cc/paper\_files/paper/2023/hash/a85b405ed65c6477a4fe8302b5e06ce7-Abstract-Conference.html} {Direct preference optimization: Your language model is secretly a reward model}.
\newblock In \emph{Advances in Neural Information Processing Systems 36: Annual Conference on Neural Information Processing Systems 2023, NeurIPS 2023, New Orleans, LA, USA, December 10 - 16, 2023}.

\bibitem[{Reimers and Gurevych(2019)}]{DBLP:conf/emnlp/ReimersG19}
Nils Reimers and Iryna Gurevych. 2019.
\newblock \href {https://doi.org/10.18653/V1/D19-1410} {Sentence-bert: Sentence embeddings using siamese bert-networks}.
\newblock In \emph{Proceedings of the 2019 Conference on Empirical Methods in Natural Language Processing and the 9th International Joint Conference on Natural Language Processing, {EMNLP-IJCNLP} 2019, Hong Kong, China, November 3-7, 2019}, pages 3980--3990. Association for Computational Linguistics.

\bibitem[{Shi et~al.(2024)Shi, Min, Yasunaga, Seo, James, Lewis, Zettlemoyer, and Yih}]{DBLP:conf/naacl/ShiMYS0LZY24}
Weijia Shi, Sewon Min, Michihiro Yasunaga, Minjoon Seo, Richard James, Mike Lewis, Luke Zettlemoyer, and Wen{-}tau Yih. 2024.
\newblock \href {https://doi.org/10.18653/V1/2024.NAACL-LONG.463} {{REPLUG:} retrieval-augmented black-box language models}.
\newblock In \emph{Proceedings of the 2024 Conference of the North American Chapter of the Association for Computational Linguistics: Human Language Technologies (Volume 1: Long Papers), {NAACL} 2024, Mexico City, Mexico, June 16-21, 2024}, pages 8371--8384. Association for Computational Linguistics.

\bibitem[{Song et~al.(2020)Song, Tan, Qin, Lu, and Liu}]{DBLP:conf/nips/Song0QLL20}
Kaitao Song, Xu~Tan, Tao Qin, Jianfeng Lu, and Tie{-}Yan Liu. 2020.
\newblock \href {https://proceedings.neurips.cc/paper/2020/hash/c3a690be93aa602ee2dc0ccab5b7b67e-Abstract.html} {Mpnet: Masked and permuted pre-training for language understanding}.
\newblock In \emph{Advances in Neural Information Processing Systems 33: Annual Conference on Neural Information Processing Systems 2020, NeurIPS 2020, December 6-12, 2020, virtual}.

\bibitem[{Springer et~al.(2024)Springer, Kotha, Fried, Neubig, and Raghunathan}]{DBLP:journals/corr/abs-2402-15449}
Jacob~Mitchell Springer, Suhas Kotha, Daniel Fried, Graham Neubig, and Aditi Raghunathan. 2024.
\newblock \href {https://doi.org/10.48550/ARXIV.2402.15449} {Repetition improves language model embeddings}.
\newblock \emph{CoRR}, abs/2402.15449.

\bibitem[{Su et~al.(2023)Su, Shi, Kasai, Wang, Hu, Ostendorf, Yih, Smith, Zettlemoyer, and Yu}]{DBLP:conf/acl/SuSKWHOYSZ023}
Hongjin Su, Weijia Shi, Jungo Kasai, Yizhong Wang, Yushi Hu, Mari Ostendorf, Wen{-}tau Yih, Noah~A. Smith, Luke Zettlemoyer, and Tao Yu. 2023.
\newblock \href {https://doi.org/10.18653/V1/2023.FINDINGS-ACL.71} {One embedder, any task: Instruction-finetuned text embeddings}.
\newblock In \emph{Findings of the Association for Computational Linguistics: {ACL} 2023, Toronto, Canada, July 9-14, 2023}, pages 1102--1121. Association for Computational Linguistics.

\bibitem[{Thirukovalluru and Dhingra(2024)}]{DBLP:journals/corr/abs-2410-14635}
Raghuveer Thirukovalluru and Bhuwan Dhingra. 2024.
\newblock \href {https://doi.org/10.48550/ARXIV.2410.14635} {Geneol: Harnessing the generative power of llms for training-free sentence embeddings}.
\newblock \emph{CoRR}, abs/2410.14635.

\bibitem[{van~den Oord et~al.(2018)van~den Oord, Li, and Vinyals}]{DBLP:journals/corr/abs-1807-03748}
A{\"{a}}ron van~den Oord, Yazhe Li, and Oriol Vinyals. 2018.
\newblock \href {https://arxiv.org/abs/1807.03748} {Representation learning with contrastive predictive coding}.
\newblock \emph{CoRR}, abs/1807.03748.

\bibitem[{Wang et~al.(2024{\natexlab{a}})Wang, Yang, Huang, Yang, Majumder, and Wei}]{DBLP:conf/acl/WangYHYMW24}
Liang Wang, Nan Yang, Xiaolong Huang, Linjun Yang, Rangan Majumder, and Furu Wei. 2024{\natexlab{a}}.
\newblock \href {https://doi.org/10.18653/V1/2024.ACL-LONG.642} {Improving text embeddings with large language models}.
\newblock In \emph{Proceedings of the 62nd Annual Meeting of the Association for Computational Linguistics (Volume 1: Long Papers), {ACL} 2024, Bangkok, Thailand, August 11-16, 2024}, pages 11897--11916. Association for Computational Linguistics.

\bibitem[{Wang and Isola(2020)}]{DBLP:conf/icml/0001I20}
Tongzhou Wang and Phillip Isola. 2020.
\newblock \href {http://proceedings.mlr.press/v119/wang20k.html} {Understanding contrastive representation learning through alignment and uniformity on the hypersphere}.
\newblock In \emph{Proceedings of the 37th International Conference on Machine Learning, {ICML} 2020, 13-18 July 2020, Virtual Event}, volume 119 of \emph{Proceedings of Machine Learning Research}, pages 9929--9939. {PMLR}.

\bibitem[{Wang et~al.(2024{\natexlab{b}})Wang, Zhao, and Lawryshyn}]{wang-etal-2024-gpt}
Yining Wang, Jinman Zhao, and Yuri Lawryshyn. 2024{\natexlab{b}}.
\newblock \href {https://aclanthology.org/2024.finnlp-2.4/} {{GPT}-signal: Generative {AI} for semi-automated feature engineering in the alpha research process}.
\newblock In \emph{Proceedings of the Eighth Financial Technology and Natural Language Processing and the 1st Agent AI for Scenario Planning}, pages 42--53, Jeju, South Korea. -.

\bibitem[{Wang et~al.(2024{\natexlab{c}})Wang, Bi, Pentyala, Ramnath, Chaudhuri, Mehrotra, Zhu, Mao, Asur, and Cheng}]{DBLP:journals/corr/abs-2407-16216}
Zhichao Wang, Bin Bi, Shiva~Kumar Pentyala, Kiran Ramnath, Sougata Chaudhuri, Shubham Mehrotra, Zixu~James Zhu, Xiang{-}Bo Mao, Sitaram Asur, and Na~Claire Cheng. 2024{\natexlab{c}}.
\newblock \href {https://doi.org/10.48550/ARXIV.2407.16216} {A comprehensive survey of {LLM} alignment techniques: Rlhf, rlaif, ppo, {DPO} and more}.
\newblock \emph{CoRR}, abs/2407.16216.

\bibitem[{Wu and Zhao(2022)}]{DBLP:conf/emnlp/WuZ22}
Bohong Wu and Hai Zhao. 2022.
\newblock \href {https://doi.org/10.18653/V1/2022.EMNLP-MAIN.221} {Sentence representation learning with generative objective rather than contrastive objective}.
\newblock In \emph{Proceedings of the 2022 Conference on Empirical Methods in Natural Language Processing, {EMNLP} 2022, Abu Dhabi, United Arab Emirates, December 7-11, 2022}, pages 3356--3368. Association for Computational Linguistics.

\bibitem[{Xia et~al.(2015)Xia, Xu, Lan, Guo, and Cheng}]{DBLP:conf/sigir/XiaXLGC15}
Long Xia, Jun Xu, Yanyan Lan, Jiafeng Guo, and Xueqi Cheng. 2015.
\newblock \href {https://doi.org/10.1145/2766462.2767710} {Learning maximal marginal relevance model via directly optimizing diversity evaluation measures}.
\newblock In \emph{Proceedings of the 38th International {ACM} {SIGIR} Conference on Research and Development in Information Retrieval, Santiago, Chile, August 9-13, 2015}, pages 113--122. {ACM}.

\bibitem[{Xiao et~al.(2022)Xiao, Liu, Shao, and Cao}]{DBLP:conf/emnlp/XiaoLSC22}
Shitao Xiao, Zheng Liu, Yingxia Shao, and Zhao Cao. 2022.
\newblock \href {https://doi.org/10.18653/V1/2022.EMNLP-MAIN.35} {Retromae: Pre-training retrieval-oriented language models via masked auto-encoder}.
\newblock In \emph{Proceedings of the 2022 Conference on Empirical Methods in Natural Language Processing, {EMNLP} 2022, Abu Dhabi, United Arab Emirates, December 7-11, 2022}, pages 538--548. Association for Computational Linguistics.

\bibitem[{Xu et~al.(2025)Xu, Pang, Shen, and Cheng}]{DBLP:conf/iclr/XuPSC25}
Shicheng Xu, Liang Pang, Huawei Shen, and Xueqi Cheng. 2025.
\newblock \href {https://openreview.net/forum?id=tbx3u2oZAu} {A theory for token-level harmonization in retrieval-augmented generation}.
\newblock In \emph{The Thirteenth International Conference on Learning Representations, {ICLR} 2025, Singapore, April 24-28, 2025}. OpenReview.net.

\bibitem[{Xu et~al.(2024{\natexlab{a}})Xu, Pang, Shen, Cheng, and Chua}]{DBLP:conf/www/XuPSCC24}
Shicheng Xu, Liang Pang, Huawei Shen, Xueqi Cheng, and Tat{-}Seng Chua. 2024{\natexlab{a}}.
\newblock \href {https://doi.org/10.1145/3589334.3645363} {Search-in-the-chain: Interactively enhancing large language models with search for knowledge-intensive tasks}.
\newblock In \emph{Proceedings of the {ACM} on Web Conference 2024, {WWW} 2024, Singapore, May 13-17, 2024}, pages 1362--1373. {ACM}.

\bibitem[{Xu et~al.(2024{\natexlab{b}})Xu, Pang, Yu, Meng, Shen, Cheng, and Zhou}]{DBLP:conf/acl/XuPYMSCZ24}
Shicheng Xu, Liang Pang, Mo~Yu, Fandong Meng, Huawei Shen, Xueqi Cheng, and Jie Zhou. 2024{\natexlab{b}}.
\newblock \href {https://doi.org/10.18653/V1/2024.ACL-LONG.9} {Unsupervised information refinement training of large language models for retrieval-augmented generation}.
\newblock In \emph{Proceedings of the 62nd Annual Meeting of the Association for Computational Linguistics (Volume 1: Long Papers), {ACL} 2024, Bangkok, Thailand, August 11-16, 2024}, pages 133--145. Association for Computational Linguistics.

\bibitem[{Zhao et~al.(2025)Zhao, Qian, Cao, Wang, Ding, Hu, Zhang, and Jin}]{zhao2025roleplayparadoxlargelanguage}
Jinman Zhao, Zifan Qian, Linbo Cao, Yining Wang, Yitian Ding, Yulan Hu, Zeyu Zhang, and Zeyong Jin. 2025.
\newblock \href {https://arxiv.org/abs/2409.13979} {Role-play paradox in large language models: Reasoning performance gains and ethical dilemmas}.
\newblock \emph{Preprint}, arXiv:2409.13979.

\bibitem[{Zhao and Zhang(2024)}]{zhao2024large}
Jinman Zhao and Xueyan Zhang. 2024.
\newblock \href {https://openreview.net/forum?id=wLQ3I0F1oj} {Large language model is not a (multilingual) compositional relation reasoner}.
\newblock In \emph{First Conference on Language Modeling}.

\bibitem[{Zhuang et~al.(2024)Zhuang, Ma, Koopman, Lin, and Zuccon}]{DBLP:conf/emnlp/ZhuangMKLZ24}
Shengyao Zhuang, Xueguang Ma, Bevan Koopman, Jimmy Lin, and Guido Zuccon. 2024.
\newblock \href {https://aclanthology.org/2024.emnlp-main.250} {Promptreps: Prompting large language models to generate dense and sparse representations for zero-shot document retrieval}.
\newblock In \emph{Proceedings of the 2024 Conference on Empirical Methods in Natural Language Processing, {EMNLP} 2024, Miami, FL, USA, November 12-16, 2024}, pages 4375--4391. Association for Computational Linguistics.

\end{thebibliography}
\newpage
\appendix

\section{Implementation details}
\label{app:details}
\paragraph{AutoRegEmbed} For the information compression task, we set the learning rate to 2e-5, the batch size to 32, and train for 2 epoch. To represent the semantics of the input, we use 5 compressed tokens. For the conditional distribution alignment task, the learning rate is set to 5e-6, with a batch size of 32 and 4 epochs. The temperature parameters $\tau$ and $\beta$ are set to 0.05 and 0.1. For the above two tasks, we set the maximum token length of context, instruction, and target to 512. Furthermore, we employ the bfloat16 format, enable FlashAttention 2, and train on four A100-80G GPUs with DeepSpeed and Zero-2. The information compression task takes 20 minutes, while the conditional distribution alignment task, involving 50,000 samples, takes approximately 1 hour. In theory, our approach eliminates the need for in-batch negative sampling, which substantially reduces memory usage. In traditional contrastive learning, a batch size of 64 requires 64 * 64=4096 pairwise computations. In contrast, our method requires only 64, reducing the number of pairwise comparisons per step by a factor of 64 and lowering memory consumption accordingly.

\paragraph{Fair Comparative Learning Baselines}
We train our own fair contrastive learning baseline based on the standard InfoNCE loss, with some code available in the FlagEmbedding repository\footnote{\url{https://github.com/FlagOpen/FlagEmbedding}}). For baselines utilizing the in-batch negative sample strategy (LLaMA2-inbatch-L and LLaMA2-inbatch-M), we experimented with batch sizes of 128, 256, 512, and 1024, determining that 512 yields the best performance. Additionally, we ensure that gradients are propagated across different devices. For baselines that do not use the in-batch negative sample strategy, we set the batch size to 32, maintaining consistency with AutoRegEmbed. Regarding the learning rate, we tested 1e-5, 5e-5, 1e-4, and 2e-4, finding that 1e-4 delivers the best results. All training data is consistent with AutoRegEmbed. We train the fair contrastive learning baseline using DeepSpeed and Zero-2 on four A100-80G GPUs in 1 hour.
\section{Variants of Equation~\ref{eq7}}
\label{app:ablation}
This section explores various possible modifications and extensions of Equation~\ref{eq7}.
\paragraph{Log\_sigmoid} Given that most loss functions are logarithmic in nature, we can modify the similarity function in Equation~\ref{eq7} by replacing the sigmoid with a Log-Sigmoid function, resulting in a more interpretable formulation:
\begin{equation*}
\begin{aligned}
\label{eq8}
&\mathcal{L}_{\mathrm{CDA}}=\mathbb{E}[-\mathrm{log}\frac{e^{S_1(q,d^+)/\tau}}{e^{S_1(q,d^+)/\tau}+\sum_i e^{S_2(d^+,d^-_i;q)/\tau}}], \\
& S_1(q,d^+) = - \mathrm{log}\sigma(\beta \ |\mathrm{log} \frac{p_{\theta_{E}}(d^+|e_{q,I_{\mathrm{next}}})}{p_{\theta_{E}}(d^+|e_{d^+,I_{\mathrm{self}}})}|  ),\\
& S_2(d^+,d^-_i;q) = 
- \mathrm{log}\sigma(\beta \ \mathrm{log}\frac{p_{\theta_{E}}(d^+|e_{q,I_{\mathrm{next}}})}{p_{\mathrm{ref}}(d^+|e_{q,I_{\mathrm{next}}})}\\
&-\beta \ \mathrm{log}\frac{p_{\theta_{E}}(d^-_i|e_{q,I_{\mathrm{next}}})}{p_{\mathrm{ref}}(d^-_i|e_{q,I_{\mathrm{next}}})}).
\end{aligned}
\end{equation*}
\paragraph{KL divergence} We also experimented with replacing the difference in log probabilities with the KL divergence between the conditional probability distributions:
\begin{equation*}
\begin{aligned}
\label{eq9}
&\mathcal{L}_{\mathrm{CDA}}=\mathbb{E}[-\mathrm{log}\frac{e^{S_1(q,d^+)/\tau}}{e^{S_1(q,d^+)/\tau}+\sum_i e^{S_2(q,d^-_i)/\tau}}], \\
& S_1(q,d^+) = - \sigma(\frac{1}{T}\sum_{t=1}^T \mathrm{KL}(p_{\theta_{E}}(d^+_t|d^+_{<t},e_{d^+,I_{\mathrm{self}}}),\\
&p_{\theta_{E}}(d^+_t|d^+_{<t},e_{q,I_{\mathrm{next}}}))),\\
& S_2(q,d^-_i) = 
- \sigma(\frac{1}{T}\sum_{t=1}^T \mathrm{KL}(p_{\theta_{E}}(d^+_t|d^-_{i,<t},e_{d^-_i,I_{\mathrm{self}}}),\\
&p_{\theta_{E}}(d^+_t|d^+_{<t},e_{q,I_{\mathrm{next}}}))).\\
\end{aligned}
\end{equation*}
\paragraph{JS divergence} In addition to KL divergence, we also employed JS divergence as a measure of distribution distance:
\begin{equation*}
\begin{aligned}
\label{eq10}
&\mathcal{L}_{\mathrm{CDA}}=\mathbb{E}[-\mathrm{log}\frac{e^{S_1(q,d^+)/\tau}}{e^{S_1(q,d^+)/\tau}+\sum_i e^{S_2(q,d^-_i)/\tau}}], \\
& S_1(q,d^+) = - \sigma(\frac{1}{T}\sum_{t=1}^T \mathrm{JS}(p_{\theta_{E}}(d^+_t|d^+_{<t},e_{d^+,I_{\mathrm{self}}}),\\
&p_{\theta_{E}}(d^+_t|d^+_{<t},e_{q,I_{\mathrm{next}}}))),\\
& S_2(q,d^-_i) = 
- \sigma(\frac{1}{T}\sum_{t=1}^T \mathrm{JS}(p_{\theta_{E}}(d^+_t|d^-_{i,<t},e_{d^-_i,I_{\mathrm{self}}}),\\
&p_{\theta_{E}}(d^+_t|d^+_{<t},e_{q,I_{\mathrm{next}}}))).\\
\end{aligned}
\end{equation*}

\paragraph{Why do KL divergence and JS divergence fail to achieve good performance in this setting?} We hypothesize that the effectiveness of using log-odds ratios stems from its ability to directly supervise the generation probability of a specific label token, thus providing a strong and targeted learning signal.
In contrast, KL and JS divergence operate over the entire output distribution across the vocabulary. While these measures are theoretically well-founded for capturing distributional differences, they often suffer from instability in practice—especially in models like LLaMA2, which have large vocabularies (~30,000 tokens). This instability weakens the gradient signal and ultimately degrades performance.
\begin{table*}[t]
\centering
\normalsize
\setlength{\tabcolsep}{3pt}
\resizebox{\linewidth}{!}{%
\begin{tabular}{lrccccccccccc}
\toprule
\textbf{Method} & \textbf{BIOSSES} &\textbf{SICK-R}&\textbf{STS12} & \textbf{STS13} & \textbf{STS14} & \textbf{STS15} & \textbf{STS16} &\textbf{STS17}&\textbf{STS22}& \textbf{STS-B}  & \textbf{Avg.}\\
\midrule
AutoRegEmbed-LLaMA2(Tanh)  & 83.97 & 80.75 & 80.58 & 86.92 & 83.19 & 88.98 & 86.96 & 85.80 & 65.91 & 85.98 & 84.77\textsubscript{(7)} / 82.90\textsubscript{(10)} \\
AutoRegEmbed-LLaMA2(Sigmoid) & 85.50 & 79.07 & 79.57 & 86.90 & 83.28 & 88.45 & 86.57 & 88.61 & 66.16 & 86.59 & 84.35\textsubscript{(7)} / 83.07\textsubscript{(10)} \\
\bottomrule
\end{tabular}
}
\caption{Performance comparison of different $\sigma$ functions for AutoRegEmbed under the same settings.}
\label{tab:analysis_sigma}
\end{table*}

\begin{table}[ht]
\centering
\small
\renewcommand{\arraystretch}{1.3}
\resizebox{\linewidth}{!}{%
\begin{tabular}{ll}
\toprule
\multicolumn{2}{l}{\textit{Retrieval Task}} \\
\midrule
$I_{\mathrm{next}}$ & \makecell[l]{Use one word to represent the query in a \\ retrieval task. The word is: ``} \\
$I_{\mathrm{self}}$ & \makecell[l]{Use one word to represent the passage in a \\ retrieval task. The word is: ``} \\
\midrule
\multicolumn{2}{l}{\textit{STS Task}} \\
\midrule
$I_{\mathrm{next}}$ & This sentence means in one word: `` \\
$I_{\mathrm{self}}$ & This sentence means in one word: `` \\
\bottomrule
\end{tabular}
}
\caption{Instructions used for Retrieval and STS tasks.}
\label{tab:instructions}
\end{table}

\section{Selection Analysis of the \texorpdfstring{$\sigma$}{sigma}}
\label{app:analysis_sigma}

To stabilize exponential operations, we apply the $\sigma$ function to map distribution-based distances to a bounded range. Sigmoid and Tanh are typical choices. Table~\ref{tab:analysis_sigma} presents their comparative performance under uniform settings.

Overall, under identical data and training settings, the Tanh variant slightly outperforms Sigmoid on 7 STS datasets, while Sigmoid shows a marginal advantage across all 10 datasets. Given the minimal performance gap, either function is a reasonable choice.

\section{Explanation of \texorpdfstring{$I_{\mathrm{next}}$ and $I_{\mathrm{self}}$}{I_next and I_self}}
\label{app:instrucitons}
Embedding tasks can be broadly categorized into sentence-to-sentence (STS) and sentence-to-document (retrieval), each associated with distinct instructions, $I_{\mathrm{next}}$ and $I_{\mathrm{self}}$. Based on prior studies \citep{DBLP:conf/emnlp/JiangHLWZ24}, the instructions used in this work are summarized in Table~\ref{tab:instructions}. Due to the symmetric nature of STS, both instructions are the same.

\section{Analysis of Alignment Strategies for Conditional Probability Distributions}
\label{app:analysis_alignment}
The motivation behind our naive alignment strategy is as follows: $I_{\mathrm{next}}$ aims for $q$ to generate $d$, while $I_{\mathrm{self}}$ guides $d$ to generate itself. Therefore, we align the probability distribution of $q$ generating $d^+$ with that of $d^+$ generating itself. Intuitively, when the conditional distributions produced by their embeddings $e_{q,I_{\mathrm{next}}}$ and $e_{d^+,I_{\mathrm{self}}}$ are closely matched, their embeddings are likely to be similar as well. Additionally, inspired by Direct Preference Optimization (DPO), we introduce negative sample feedback by encouraging $e_{q,I_{\mathrm{next}}}$ to increase the distance to negative samples while decreasing the distance to positive ones.

However, the use of instructions enables more sophisticated strategies for aligning the probability distributions between $q$ and $d$. To explore this, we conducted a comprehensive and detailed analysis. We begin by introducing several alignment strategy configurations.

\paragraph{Strategy 1} In this strategy, we align the conditional probability distribution of generating $q$ rather than $d^+$. Specifically, we adjust the instructions to treat $q$ as the anchor, aligning the probability distribution between $q$ and $d^+$. This is implemented by simply replacing the corresponding prompts, as detailed in Equation~\ref{eq11}.
For the STS task, where symmetry holds, $I_{\mathrm{self}}$ and $I_{\mathrm{prev}}$ are identical.
For the asymmetric retrieval task, an example of $I_{\mathrm{prev}}$ could be:
“Use a word to express the question that this article can answer:``”. 
\begin{equation}
\begin{aligned}
\label{eq11}
&S_1(q,d^+) = - \sigma(\beta \ |\mathrm{log} \frac{p_{\theta_{E}}(q|e_{q,I_{\mathrm{self}}})}{p_{\theta_{E}}(q|e_{d^+,I_{\mathrm{prev}}})}|  )\\
& S_2(d^+,d^-_i;q) = 
- \sigma(\beta \ \mathrm{log}\frac{p_{\theta_{E}}(q|e_{d^+,I_{\mathrm{prev}}})}{p_{\mathrm{ref}}(q|e_{d^+,I_{\mathrm{prev}}})}\\
&-\beta \ \mathrm{log}\frac{p_{\theta_{E}}(q|e_{d^-_i,I_{\mathrm{prev}}})}{p_{\mathrm{ref}}(q|e_{d^-_i,I_{\mathrm{prev}}})}).
\end{aligned}
\end{equation}

\paragraph{Strategy 2} We treat both $q$ and $d^+$ as anchors and compute a weighted similarity between their respective conditional distributions.
\begin{equation}
\begin{aligned}
\label{eq12}
&S_1(q,d^+) = - \frac{1}{2}\sigma(\beta \ |\mathrm{log} \frac{p_{\theta_{E}}(q|e_{q,I_{\mathrm{self}}})}{p_{\theta_{E}}(q|e_{d^+,I_{\mathrm{prev}}})}|  ) \\ 
&-\frac{1}{2}\sigma(\beta \ |\mathrm{log} \frac{p_{\theta_{E}}(d^+|e_{q,I_{\mathrm{next}}})}{p_{\theta_{E}}(d^+|e_{d^+,I_{\mathrm{self}}})}|  )\\
& S_2(d^+,d^-_i;q) = - \frac{1}{2}\sigma(\beta \ \mathrm{log}\frac{p_{\theta_{E}}(d^+|e_{q,I_{\mathrm{next}}})}{p_{\mathrm{ref}}(d^+|e_{q,I_{\mathrm{next}}})}\\
&-\beta \ \mathrm{log}\frac{p_{\theta_{E}}(d^-_i|e_{q,I_{\mathrm{next}}})}{p_{\mathrm{ref}}(d^-_i|e_{q,I_{\mathrm{next}}})})\\
&- \frac{1}{2}\sigma(\beta \ \mathrm{log}\frac{p_{\theta_{E}}(q|e_{d^+,I_{\mathrm{prev}}})}{p_{\mathrm{ref}}(q|e_{d^+,I_{\mathrm{prev}}})}\\
&-\beta \ \mathrm{log}\frac{p_{\theta_{E}}(q|e_{d^-_i,I_{\mathrm{prev}}})}{p_{\mathrm{ref}}(q|e_{d^-_i,I_{\mathrm{prev}}})}).
\end{aligned}
\end{equation}

\begin{table*}[ht]
\centering
\small
\setlength{\tabcolsep}{4pt}
\renewcommand{\arraystretch}{1.2}
\resizebox{\linewidth}{!}{%
\begin{tabular}{l c c c c c c c c c c c c}
\toprule
\textbf{Method} & \textbf{Params} & \textbf{BIOSSES} & \textbf{SICK-R} & \textbf{STS12} & \textbf{STS13} & \textbf{STS14} & \textbf{STS15} & \textbf{STS16} & \textbf{STS17} & \textbf{STS22} & \textbf{STS-B} & \textbf{Avg.} \\
\midrule
Strategy 1 & 7B & 86.31 & 81.97 & 81.05 & 86.27 & 82.80 & 88.51 & 86.31 & 86.43 & 64.23 & 86.35 & 84.75\textsubscript{(7)} / 83.02\textsubscript{(10)} \\
Strategy 2 & 7B & 83.34 & 81.87 & 79.82 & 84.11 & 81.97 & 89.11 & 86.39 & 88.79 & 66.11 & 86.90 & 84.32\textsubscript{(7)} / 82.84\textsubscript{(10)} \\
Strategy 3 & 7B & 86.03 & 83.20 & 78.35 & 81.54 & 81.27 & 87.80 & 87.04 & 89.16 & 64.55 & 86.28 & 83.64\textsubscript{(7)} / 82.52\textsubscript{(10)} \\
Strategy 4 & 7B & 84.10 & 80.74 & 78.09 & 83.79 & 81.77 & 88.01 & 86.14 & 89.64 & 64.14 & 87.16 & 83.67\textsubscript{(7)} / 82.36\textsubscript{(10)} \\
AutoRegEmbed    & 7B & 85.50 & 79.07 & 79.57 & 86.90 & 83.28 & 88.45 & 86.57 & 88.61 & 66.16 & 86.59 & 84.35\textsubscript{(7)} / 83.07\textsubscript{(10)} \\
\bottomrule
\end{tabular}
}
\caption{Performance comparison of AutoRegEmbed under different alignment strategies on STS benchmarks.}
\label{tab:autoregembed_alignment_strategies}
\end{table*}

\paragraph{Strategy 3} We can flexibly expand the range of negative samples by varying instructions—for example, by increasing the probability of positive samples generating themselves, while decreasing the probability of them generating negative samples.

\begin{equation}
\begin{aligned}
\label{eq13}
&\mathcal{L}_{\mathrm{CDA}}=\mathbb{E}[-\mathrm{log}\frac{e^{S_1(q,d^+)/\tau}}{e^{S_1(q,d^+)/\tau}+\sum_i e^{S_2(d^+,d^-_i;q)/\tau}}], \\
& S_1(q,d^+) = - \sigma(\beta \ |\mathrm{log} \frac{p_{\theta_{E}}(d^+|e_{q,I_{\mathrm{next}}})}{p_{\theta_{E}}(d^+|e_{d^+,I_{\mathrm{self}}})}|  ),\\
& S_2(d^+,d^-_i;q) = 
- \frac{1}{2}\sigma(\beta \ \mathrm{log}\frac{p_{\theta_{E}}(d^+|e_{q,I_{\mathrm{next}}})}{p_{\mathrm{ref}}(d^+|e_{q,I_{\mathrm{next}}})}\\
&-\beta \ \mathrm{log}\frac{p_{\theta_{E}}(d^-_i|e_{q,I_{\mathrm{next}}})}{p_{\mathrm{ref}}(d^-_i|e_{q,I_{\mathrm{next}}})})\\
&- \frac{1}{2}\sigma(\beta \ \mathrm{log}\frac{p_{\theta_{E}}(d^+|e_{d^+,I_{\mathrm{self}}})}{p_{\mathrm{ref}}(d^+|e_{d^+,I_{\mathrm{self}}})}\\
&-\beta \ \mathrm{log}\frac{p_{\theta_{E}}(d^-_i|e_{d^+,I_{\mathrm{self}}})}{p_{\mathrm{ref}}(d^-_i|e_{d^+,I_{\mathrm{self}}})})
\end{aligned}
\end{equation}

\paragraph{Strategy 4} Similar to Strategy 3, we can also increase the probability of negative samples generating themselves, while reducing the probability of them generating positive samples, as shown in Equation~\ref{eq14}.
\begin{table*}[ht]
\centering
\small
\renewcommand{\arraystretch}{1.2}
\resizebox{\linewidth}{!}{%
\begin{tabular}{l c c c c c c c c c c c c}
\toprule
\textbf{Method} & \textbf{Params} & \textbf{BIOSSES} & \textbf{SICK-R} & \textbf{STS12} & \textbf{STS13} & \textbf{STS14} & \textbf{STS15} & \textbf{STS16} & \textbf{STS17} & \textbf{STS22} & \textbf{STS-B} & \textbf{Avg.} \\
\midrule
AutoRegEmbed($\tau{=}0.1, \beta{=}0.1$) & 7B & 85.50 & 79.07 & 79.57 & 86.90 & 83.28 & 88.45 & 86.57 & 88.61 & 66.16 & 86.59 & 84.35\textsubscript{(7)} / 83.07\textsubscript{(10)} \\
AutoRegEmbed($\tau{=}0.02, \beta{=}0.1$)      & 7B & 84.90 & 79.85 & 78.58 & 85.54 & 84.64 & 88.71 & 86.88 & 87.57 & 65.61 & 86.02 & 84.32\textsubscript{(7)} / 82.83\textsubscript{(10)} \\
AutoRegEmbed($\tau{=}0.05, \beta{=}0.1$)      & 7B & 84.65 & 81.46 & 79.98 & 86.35 & 83.33 & 89.21 & 86.91 & 87.67 & 65.90 & 86.98 & 84.89\textsubscript{(7)} / 83.24\textsubscript{(10)} \\
AutoRegEmbed($\tau{=}0.2, \beta{=}0.1$)       & 7B & 83.85 & 81.72 & 79.77 & 86.73 & 83.19 & 88.41 & 86.53 & 87.69 & 66.22 & 86.37 & 84.67\textsubscript{(7)} / 83.05\textsubscript{(10)} \\
AutoRegEmbed($\tau{=}1.0, \beta{=}0.1$)       & 7B & 81.23 & 80.57 & 77.63 & 83.90 & 81.92 & 87.08 & 85.75 & 88.18 & 63.91 & 85.95 & 83.26\textsubscript{(7)} / 81.61\textsubscript{(10)} \\
AutoRegEmbed($\tau{=}0.1, \beta{=}0.2$)       & 7B & 84.06 & 81.49 & 80.32 & 87.15 & 83.49 & 88.67 & 86.85 & 87.22 & 66.45 & 86.56 & 84.93\textsubscript{(7)} / 83.23\textsubscript{(10)} \\
AutoRegEmbed($\tau{=}0.1, \beta{=}0.3$)       & 7B & 84.50 & 79.56 & 79.45 & 86.62 & 83.55 & 87.78 & 86.01 & 89.77 & 65.63 & 86.13 & 84.16\textsubscript{(7)} / 82.90\textsubscript{(10)} \\
AutoRegEmbed($\tau{=}0.1, \beta{=}0.4$)       & 7B & 84.27 & 81.04 & 78.92 & 85.76 & 82.44 & 88.54 & 86.05 & 87.48 & 65.99 & 86.08 & 84.12\textsubscript{(7)} / 82.65\textsubscript{(10)} \\
AutoRegEmbed($\tau{=}0.05, \beta{=}0.2$)      & 7B & 84.31 & 79.65 & 79.59 & 84.16 & 81.95 & 89.53 & 87.54 & 89.37 & 66.78 & 87.48 & 84.27\textsubscript{(7)} / 83.04\textsubscript{(10)} \\
\bottomrule
\end{tabular}
}
\caption{Ablation of temperature $\tau$ and alignment weight $\beta$ on STS benchmarks. The Avg. column shows results over 7 and 10 datasets.}
\label{tab:temperature_ablation}
\end{table*}

Table~\ref{tab:autoregembed_alignment_strategies} presents the performance of the LLaMA2-7B model trained on the STS split of the MEDI dataset under various alignment strategies. Both $\tau$ and $\beta$ are fixed at 0.1.

\begin{equation}
\begin{aligned}
\label{eq14}
&\mathcal{L}_{\mathrm{CDA}}=\mathbb{E}[-\mathrm{log}\frac{e^{S_1(q,d^+)/\tau}}{e^{S_1(q,d^+)/\tau}+\sum_i e^{S_2(d^+,d^-_i;q)/\tau}}], \\
& S_1(q,d^+) = - \sigma(\beta \ |\mathrm{log} \frac{p_{\theta_{E}}(d^+|e_{q,I_{\mathrm{next}}})}{p_{\theta_{E}}(d^+|e_{d^+,I_{\mathrm{self}}})}|  ),\\
& S_2(d^+,d^-_i;q) = 
- \frac{1}{2}\sigma(\beta \ \mathrm{log}\frac{p_{\theta_{E}}(d^+|e_{q,I_{\mathrm{next}}})}{p_{\mathrm{ref}}(d^+|e_{q,I_{\mathrm{next}}})}\\
&-\beta \ \mathrm{log}\frac{p_{\theta_{E}}(d^-_i|e_{q,I_{\mathrm{next}}})}{p_{\mathrm{ref}}(d^-_i|e_{q,I_{\mathrm{next}}})})\\
&- \frac{1}{2}\sigma(\beta \ \mathrm{log}\frac{p_{\theta_{E}}(d^-_i|e_{d^-_i,I_{\mathrm{self}}})}{p_{\mathrm{ref}}(d^-|e_{d^-_i,I_{\mathrm{self}}})}\\
&-\beta \ \mathrm{log}\frac{p_{\theta_{E}}(d^+|e_{d^-_i,I_{\mathrm{self}}})}{p_{\mathrm{ref}}(d^+|e_{d^-_i,I_{\mathrm{self}}})})
\end{aligned}
\end{equation}
The experimental results reveal a consistent trend: the simpler the modification to the original loss function (Equation~\ref{eq7}), the better the performance. Specifically, \textbf{Strategy 1} alters the generation target from $d^+$ to $q$. Given the inherent symmetry between the two, this change leads to no significant performance difference. \textbf{Strategy 2} computes a weighted average of the generation probabilities for $q$ and $d^+$; however, despite its increased computational cost, it does not improve performance. \textbf{Strategies 3 and 4} introduce more complex alignment designs involving positive and negative samples, but these result in inferior performance.

\section{Analysis of Temperature Coefficients $\tau$ and $\beta$}
\label{app:analysis_temperature}
The reasons for introducing these two temperature coefficients are as follows:
\begin{itemize}
    \item We introduce the temperature coefficient $\tau$ to align our loss function more closely with the InfoNCE objective. Intuitively, $\tau$ controls the strength of separation between positive and negative samples—the lower the value, the sharper the contrast enforced by the loss.
    \item The temperature coefficient $\beta$ is inspired by the DPO (Direct Preference Optimization) loss, which uses a similar scaling factor to modulate the influence of preference differences—also expressed as log probability ratios. In our setting, $\beta$ controls the sensitivity of the loss to differences between distributions.
\end{itemize}
To empirically assess the impact of these hyperparameters, we conducted a grid search using the LLaMA2-7B model and the STS split from the MEDI dataset. For $\tau$, we adopted values commonly used in contrastive learning: ${0.02,\ 0.05,\ 0.1,\ 0.2,\ 1.0}$. For $\beta$, we selected values based on prior DPO studies: ${0.1,\ 0.2,\ 0.3,\ 0.4}$. After systematic evaluation, we identified the best-performing $(\tau, \beta)$ pair.

To empirically assess the impact of these hyperparameters, we conducted a grid search using the LLaMA2-7B model and the STS split from the MEDI dataset. For $\tau$, we adopted values commonly used in contrastive learning: ${0.02,\ 0.05,\ 0.1,\ 0.2,\ 1.0}$. For $\beta$, we selected values based on prior DPO studies: ${0.1,\ 0.2,\ 0.3,\ 0.4}$. After systematic evaluation,  we observe that the model performs better when $\tau=0.05$ or $\beta=0.2$. Furthermore, except for a few extreme cases (e.g., $\tau=1.0$ or $\beta=0.4$), the model’s performance remains relatively stable across different settings, which demonstrates the robustness of our method. When $\tau=0.05$ and $\beta=0.2$ are applied simultaneously, the model does not exhibit improved performance. This suggests that the effects of these two temperature parameters may not be orthogonal.

\end{document}